\documentclass{article}
\usepackage{graphicx}
\usepackage{amsmath}
\usepackage{amssymb}
\usepackage{placeins}
\usepackage{subcaption}[section]
\usepackage{caption}
\usepackage{xcolor}
\usepackage{hyperref}
\usepackage{booktabs}
\usepackage{cite}

\usepackage{multirow}

\usepackage{float}
\usepackage[utf8]{inputenc}
\usepackage[longend]{algorithm2e}
\usepackage{textgreek}
\usepackage{amssymb}
\usepackage{tcolorbox}

\usepackage[utf8]{inputenc}

\usepackage[preprint]{corl_2021} 

\title{Analyzing Effects of The COVID-19 Pandemic on Road Traffic Safety: The Cases of New York City, Los Angeles, and Boston}

\author{
  Lahari Karadla\\
  Department of Computer Science\\
  University of Memphis\\
  \texttt{skaradla@memphis.edu} \\
}

\begin{document}
\maketitle

\begin{abstract}
The COVID-19 pandemic has resulted in significant social and economic impacts throughout the world. In addition to the health consequences, the impacts on traffic behaviors have also been sudden and dramatic. We have analyzed how the road traffic safety of New York City, Los Angeles, and Boston in the U.S. have been impacted by the pandemic and corresponding local government orders and restrictions. To be specific, we have studied the accident hotspots' distributions before and after the outbreak of the pandemic and found that traffic accidents have shifted in both location and time compared to previous years. In addition, we have studied the road network characteristics in those hotspot regions with the hope to understand the underlying cause of the hotspot shifts.  

\end{abstract}


\section{Introduction} 

The year 2020 will be remembered for the COVID-19 pandemic, and all of its effects on several million people all over the world. One of the domino effects of the pandemic was the imposition of lockdowns to reduce the spread of infection and potential deaths, resulting in unprecedented traffic control~\cite{Wang2021Mobility,Lin2021Safety}.
According to~\cite{zhang2021effect}, in the case of COVID-19, the ``stay-at-home'' order and implications of social distancing measures have significantly changed people’s travel behavior in many ways. The percentage of people staying at home practicing social distancing contributed to change in transport shows a trend of less congestion and accidents. In this study, we evaluate traffic accident datasets from 2018 to 2021. The mobility analysis reveals significant changes of mobility patterns through identifying the potential long-term effect of the pandemic on road traffic safety.

The parameters we comprehend from this study can help understand traffic pattern changes alone with the road networks in detail. The reduction in the number of traffic accidents in the region was analysed, as were more specific details, such as territorial location, severity, distribution of the types of road on which they occurred. This research will contribute to the ongoing work on the effects of the COVID-19 on transportation around the world. In addition, our observations from studying the mobility change point, accident hotspot shift, and road network characteristics will help understand the underlying cause of the change of traffic patterns.

\section{Related Work}

We first review existing studies which analyze spatial-temporal characteristics of traffic accidents; then, we move on to consider recent work assessing the impacts of COVID-19 on road traffic safety.

\subsection{Spatial and temporal accident analysis} 

The spatial-temporal view of the accidents can help us road traffic patterns regarding traffic accidents. Studies such as~\cite{xie2008kernel} and~\cite{kang2018spatiotemporal} have aimed to identify frequent accident prone locations.~\cite{kim2007using} uses K-means clustering method to analyze spatial patterns of pedestrian involved crashed in Honolulu, Hawaii. Community detection algorithm and the association rule learning algorithm are used in~\cite{lin2014data} to identify important accident characteristics, reducing the inherent heterogeneity to discern meaning patterns.~\cite{lin2014data} puts forward a methodology to identify high density accident hotspots. The kernel density estimation enables an overarching visualisation and manipulation of the accidents based on density which is used in turn to create the basic spatial unit for the hotspot clustering method. 

The temporal point of the view of various traffic safety interventions is explored in detail by the following studies~\cite{wahl2010red,robson2001guide,missoni2012alcohol,green2016traffic,haghpanahan2019evaluation}. In~\cite{robson2001guide}, the authors explain that a significant reduction in the number of accidents occurred despite the number of cars entering the intersection during a red light is reduced.

\subsection{COVID-19's impact on road traffic safety}

The lockdown measures implemented due to COVID-19 significantly reduced domestic travel in U.S. Comparing early March and Mid April of 2020, the travel has dropped 71\% according to~\cite{trb2020}. However, a report from the National Safety Council shows a 14\% increase in the traffic accident fatality rate during March 2020 compared to March 2019 surprising everyone~\cite{NSC2020}. 
 
Along with analyzing traffic rates, various studies have modeled effects of lockdown mathematically. To provide some examples, in~\cite{barnes2020effect} the authors argue that according to data from Google Community Mobility reports and Uniform Traffic Crash Report, they find that the ``stay-at-home'' order has led to a large decrease in traffic accidents (proved using a discontinuity regression). Other similar studies have used the difference-in-differences analysis to evaluate the frequency of vehicle collisions further. 

All above-mentioned studies assume that people have followed the orders and guidelines issued by the government and adjusted their mobility patterns. There is a lack in systematic approach to detect mobility change point in existing studies which could result in an erroneous analysis of the pandemic's impact~\cite{Lin2021Safety}. In addition, the previous mentioned studies do not explore the spatial distribution of accident hotspots, which would be beneficial in designing better and effective road safety interventions.


\section{Methods}
In this section, we briefly discuss the methods used in the process of our analysis. 

\subsection{Change point detection}

Change point detection is an important part of time series analysis, as the presence of a change point indicates an abrupt and significant change in the data generating process. 
Consider a non-stationary time series $ m =\{m_{t}\}_{t=1}^{T}  $ with possible changes at $K$ unknown time steps 
$ 1 < t_{1} < t_{2} <...< t_{k} < T $. Change-point detection algorithm helps find these unknown time steps solving this optimization program:

           \[ \min\limits_{\tau} V (\tau) + {\beta}K \]               

where $\beta$ is the weighting factor;  $ \tau = \{t_1, t_2,..., t_K\}$ serves as the example to explain division of the time series. Both $\tau$ and $K$ are unknown and is determined by optimization algorithm. $V(\tau)$ is defined as:

   \[ V (\tau ) = \sum_{k=0}^{K} c(m_{t_k}...m_{t_{k+1}}) \]  

where  $t_{0} = 1 $ and $ t_{K+1} = T$. $c(·)$ in this equation is the measure of similarity of  elements in the time series segment $m_{t_k}..m_{t_k+1} = \{m_{t}\}_{t_k+1}^{t_k}.$. In this study $c(·)$ is a function to determine the extent to which the observed samples approximate to theoretical default value set according to~\cite{truong2020selective}.






\subsection{Kernel Density Estimation}

The purpose of kernel density estimation (KDE) is to produce a density surface of point events over space by computing event intensity as density estimation. To analyze the accident locations in our study, we perform KDE to construct the spatial-temporal distribution.
Take the 30-day intervals as examples, we consider accident data of every month from 2018 to 2021. Any change in patterns during the pandemic will be compared to the results of 2019.
A global two-sample test on the accident locations will help produce quantitative results with respect to the integrated squared error (ISE) between two density functions $f_1$ and $f_2$.

\[ ISE = \int ( f_1(x) - f_2(x))^2  \ dx \]

where the null hypothesis is $H0 : f_{1} = f_{2}$.

\section{Datasets}

 Google Community Mobility Reports are considered for detecting the mobility change point~\cite{Google}, which contain daily percentage changes of traffic with purposes such as retail, recreation, healthcare, grocery and more. A combined percentage of all traffic is considered as a metric to analyze the mobility trend of a particular region. For the sources of our data, we have considered vehicle crashes data from New York City, Los Angeles, and Boston~\cite{newyorkcity,Ladata,bostondata}. The road network detailed file is obtained from openstreet map website. We estimate the shift in accident hotspot and the corresponding road data is obtained. All these datasets includes date, latitude, longitude of every accident. Various demographic parameters including age, severity, race, gender are also recorded. The studying interval is considered to be 30-day before and after the detected mobility change point.

\section{Results}
The study~\cite{trafficnytimes} has reported that nearly all countries have experienced a 70\% decrease in traffic as a result of the pandemic beginning the inspiration for our study. We have considered the geographic locations of traffic accidents in New York city (NYC), Los Angeles (LA), and Boston to understand the spatial-temporal distribution of accident hotspots. We use 30-day interval traffic accident counts of NYC, LA, and Boston from 2018 to 2021, which consist of attributes such as the date of the accident, latitude, longitude, and various demographic factors to explain the data in detail.

Once this data is processed, we perform KDE on the datasets to study any impact due to change in traffic patterns. This will help provide the likelihood of an event thus modeling the uncertainty of an accident occurring at a location. According to~\cite{xie2008kernel}, from a geographic point of view,  the occurrence of traffic accidents along a certain roadway segment is largely determined by its traffic volume, which exhibits distinct spatial and temporal patterns.

\begin{figure}[H]
  \subfloat[NYC January 2018]{
	\begin{minipage}[c][1\width]{
	   0.2\textwidth}
	   \centering
	   \includegraphics[width=1.2\textwidth]{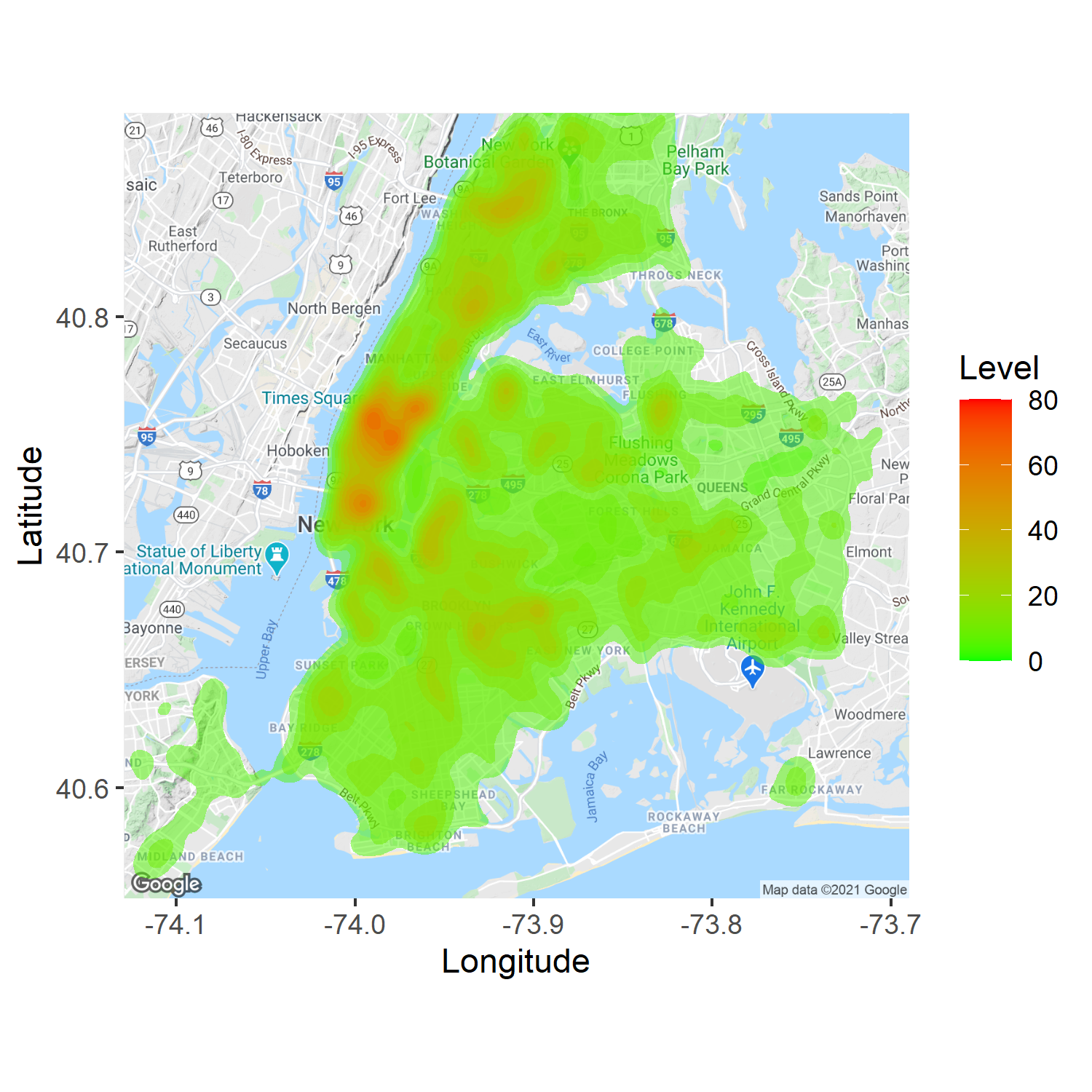}
	\end{minipage}}
 \hfill 	
  \subfloat[NYC January 2019]{
	\begin{minipage}[c][1\width]{
	   0.2\textwidth}
	   \centering
	   \includegraphics[width=1.2\textwidth]{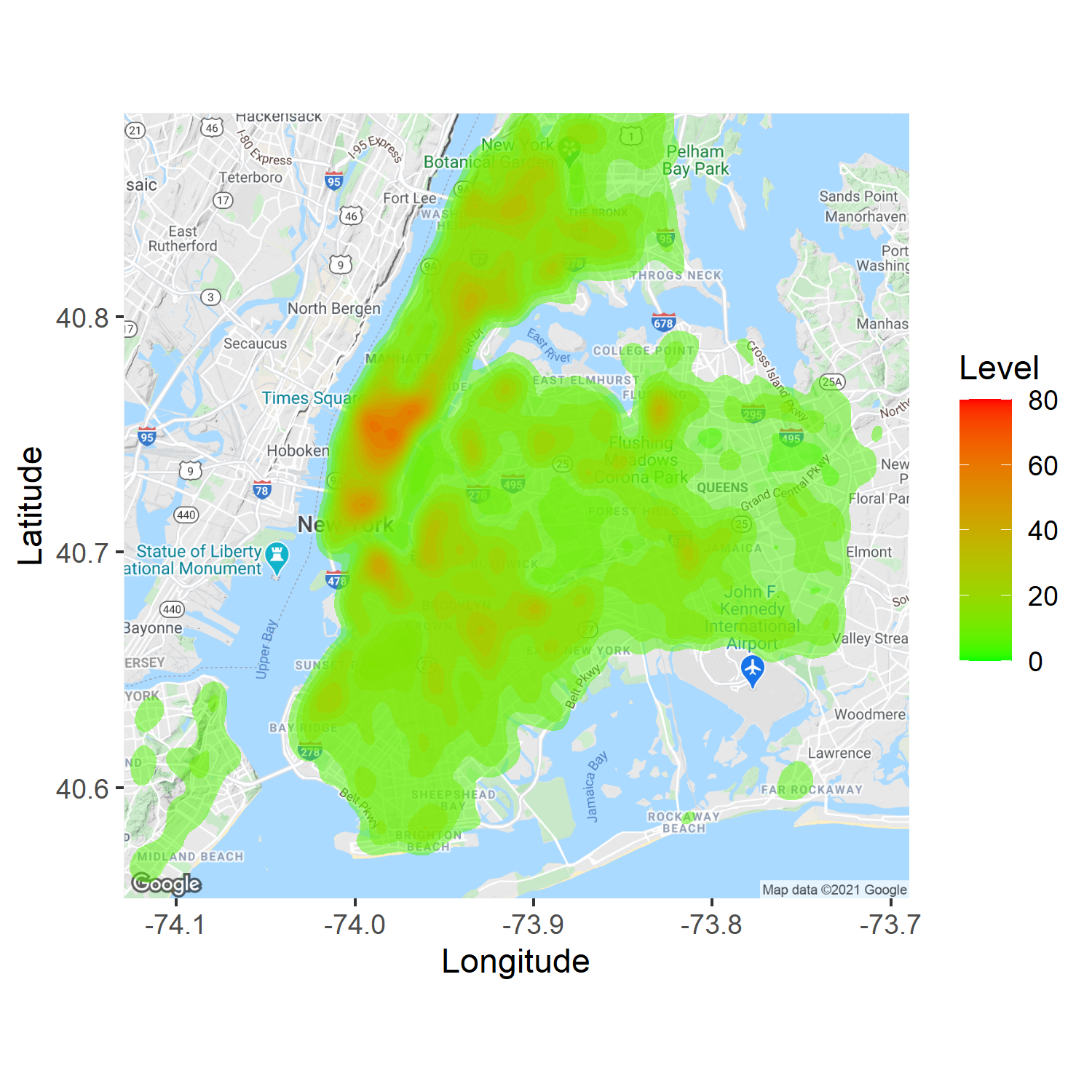}
	\end{minipage}}
 \hfill	
  \subfloat[NYC January 2020]{
	\begin{minipage}[c][1\width]{
	   0.2\textwidth}
	   \centering
	   \includegraphics[width=1.2\textwidth]{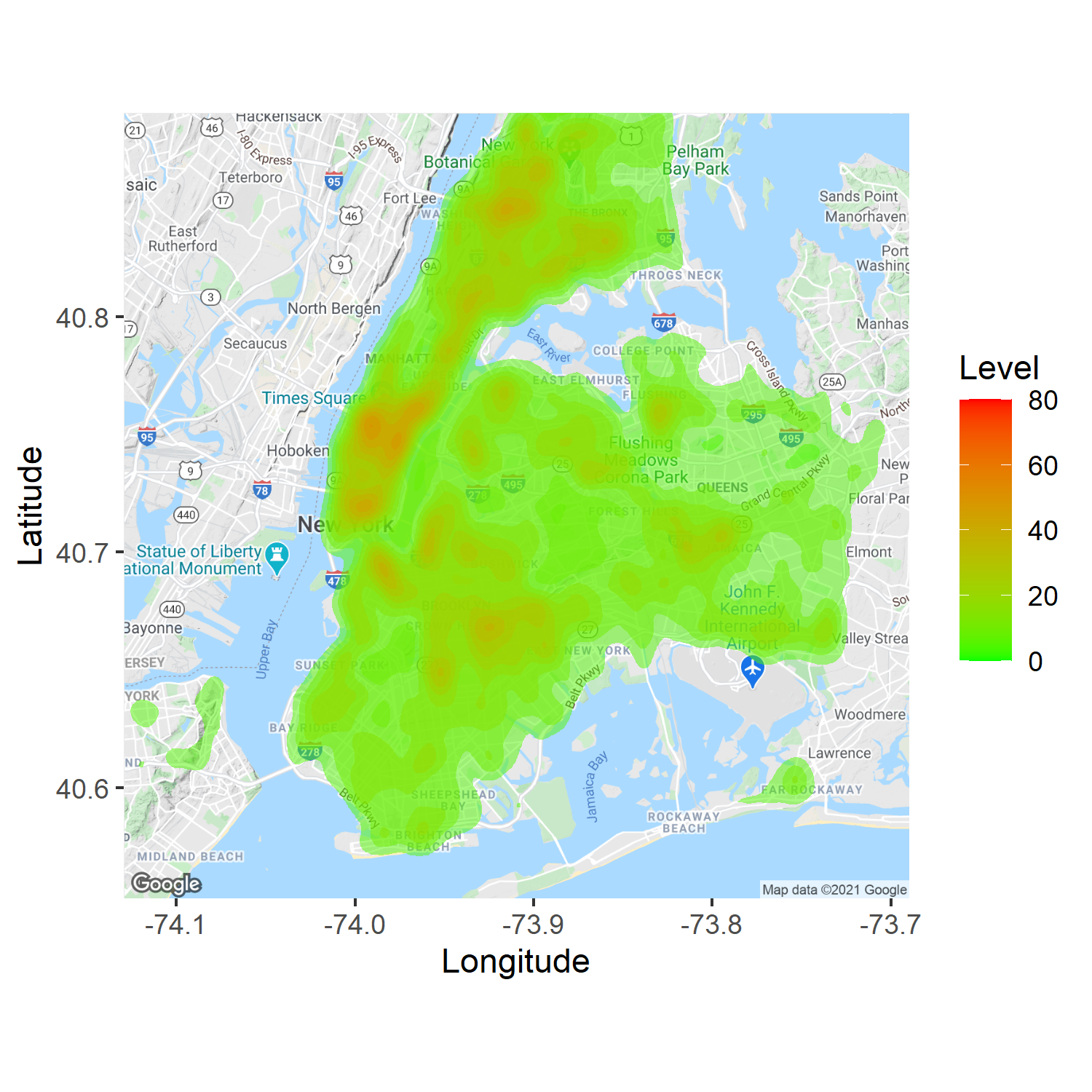}
	\end{minipage}}
 \hfill	
  \subfloat[NYC January 2021]{
	\begin{minipage}[c][1\width]{
	   0.2\textwidth}
	   \centering
	   \includegraphics[width=1.2\textwidth]{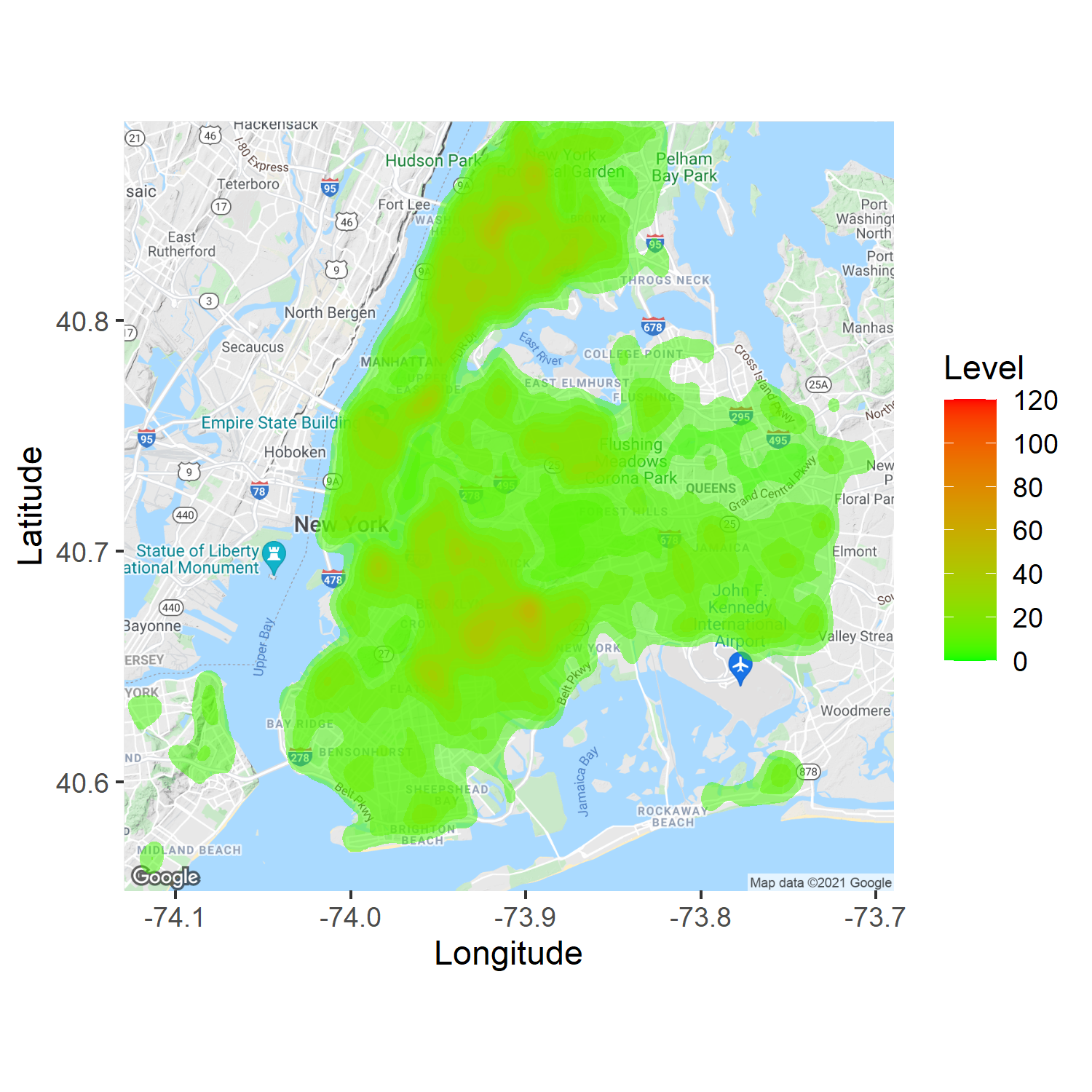}
	\end{minipage}}
\caption{The KDE results of traffic accidents in New York City. Four 30-day analysis of January, 2018--2021 are shown. The accident hotspots have shifted from Midtown and Lower Manhattan to Upper East Side, West Bronx, and southern Brooklyn.}
\label{fig:nycjan}
\end{figure}

In all three cities, the results of 2018--2021 are shown. Our first observation is in Figure~\ref{fig:nycjan} (a)--(d) and all the following results would be the difference in accident hotspots and their shift in patterns. The reduction in traffic and accidents leads to some inconsistency in the mobility change point regions. We can see that the accident hotspots have been shifted to different locations before and after the pandemic. To be specific, in NYC, the hotspots are shifted from Midtown and Lower Manhattan to the Upper East side, and West Bronx and Southern Brooklyn. In LA, the hotspots are shifted from Hollywood and northern LA to southern LA. In Boston, the hotspots are shifted to the south of the city.

\begin{figure}[H]
  \subfloat[NYC March 2018]{
	\begin{minipage}[c][1\width]{
	   0.2\textwidth}
	   \centering
	   \includegraphics[width=1.2\textwidth]{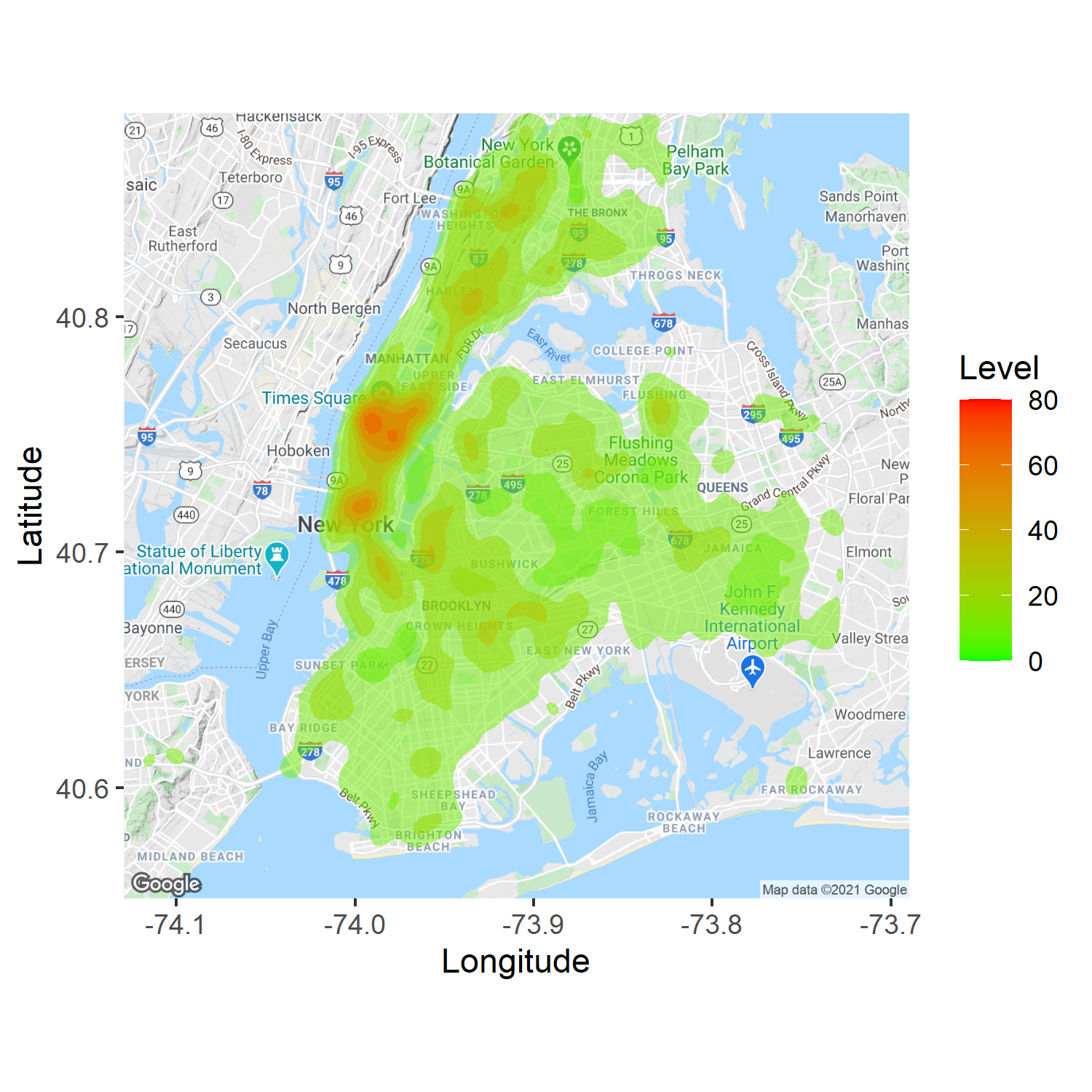}
	\end{minipage}}
 \hfill 	
  \subfloat[NYC March 2019]{
	\begin{minipage}[c][1\width]{
	   0.2\textwidth}
	   \centering
	   \includegraphics[width=1.2\textwidth]{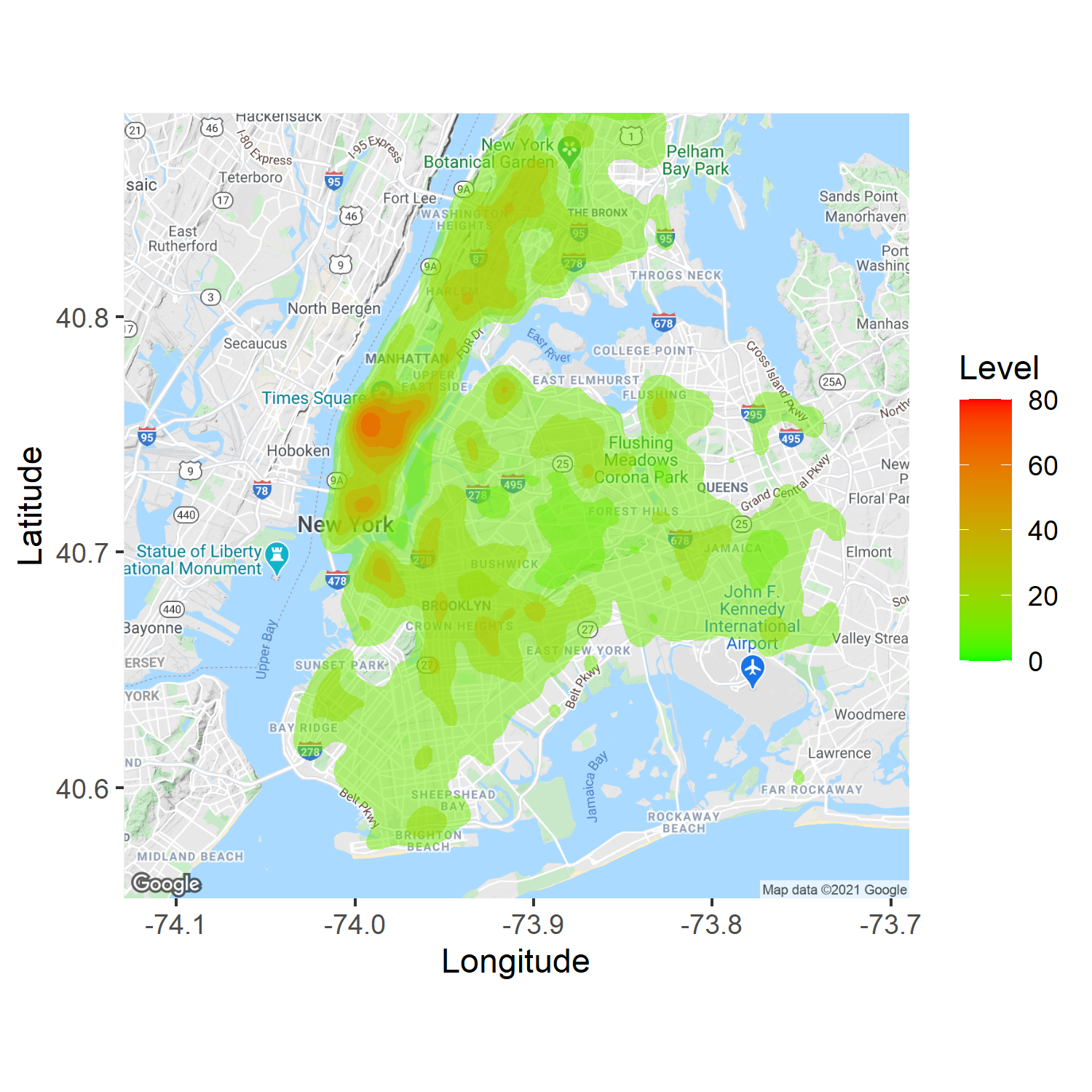}
	\end{minipage}}
 \hfill	
  \subfloat[NYC March 2020]{
	\begin{minipage}[c][1\width]{
	   0.2\textwidth}
	   \centering
	   \includegraphics[width=1.2\textwidth]{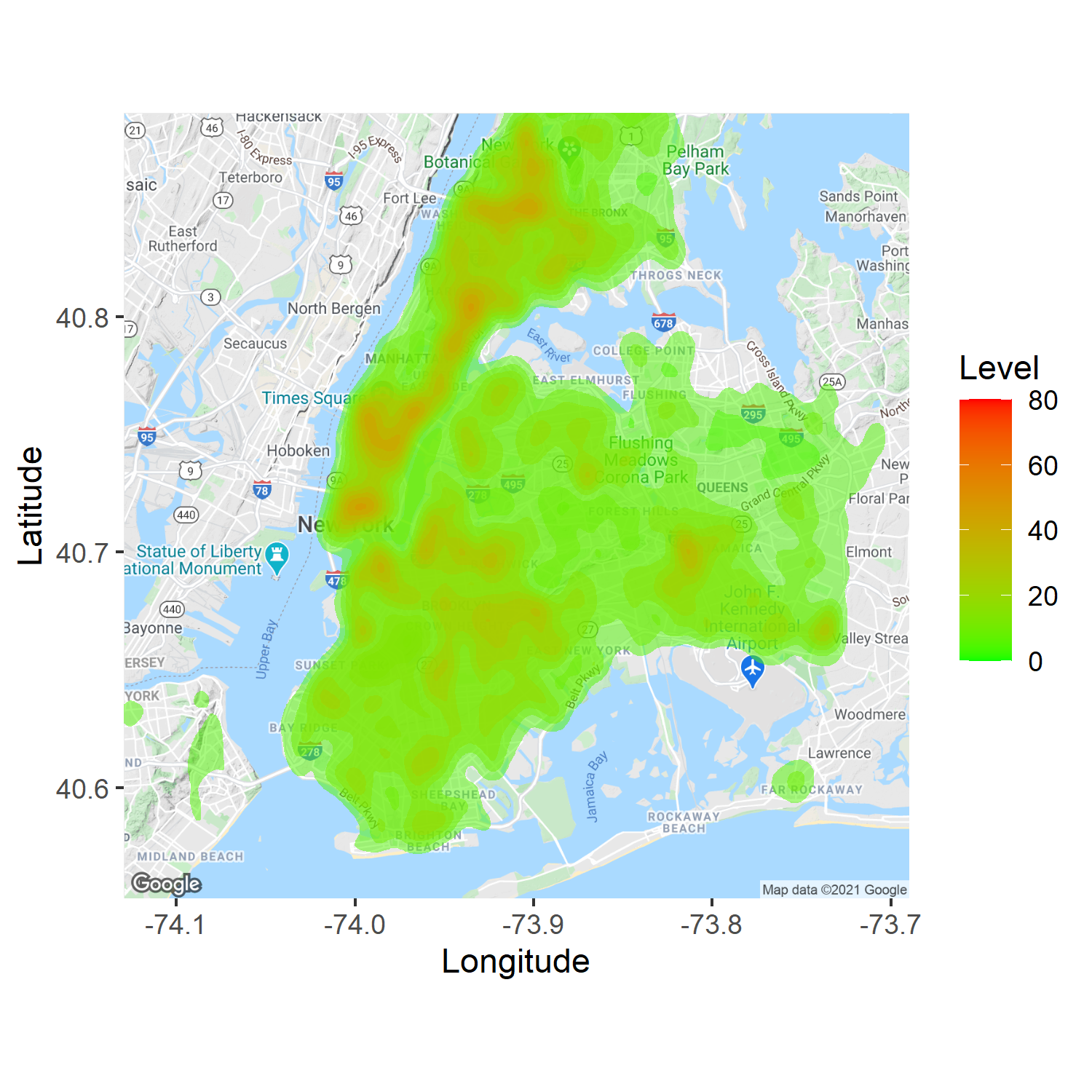}
	\end{minipage}}
 \hfill	
  \subfloat[NYC March 2021]{
	\begin{minipage}[c][1\width]{
	   0.2\textwidth}
	   \centering
	   \includegraphics[width=1.2\textwidth]{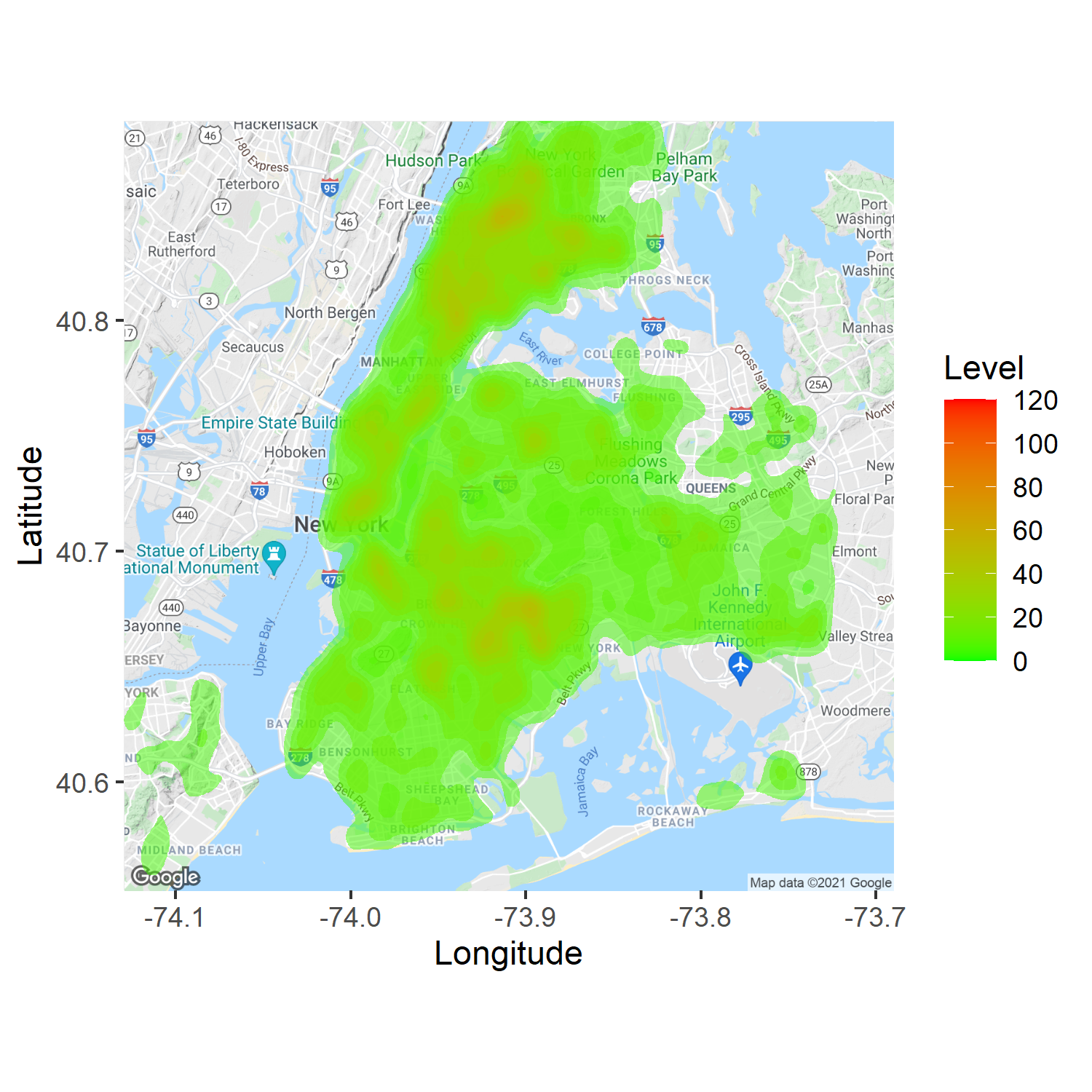}
	\end{minipage}}
\caption{The KDE results of traffic accidents in New York City. Four 30-day analysis of March, 2018--2021 are shown. The accident hotspots have shifted from Midtown and Lower Manhattan to Upper East Side, West Bronx, and southern Brooklyn.}
\end{figure}

\begin{figure}[H]
  \subfloat[NYC June 2018]{
	\begin{minipage}[c][1\width]{
	   0.2\textwidth}
	   \centering
	   \includegraphics[width=1.2\textwidth]{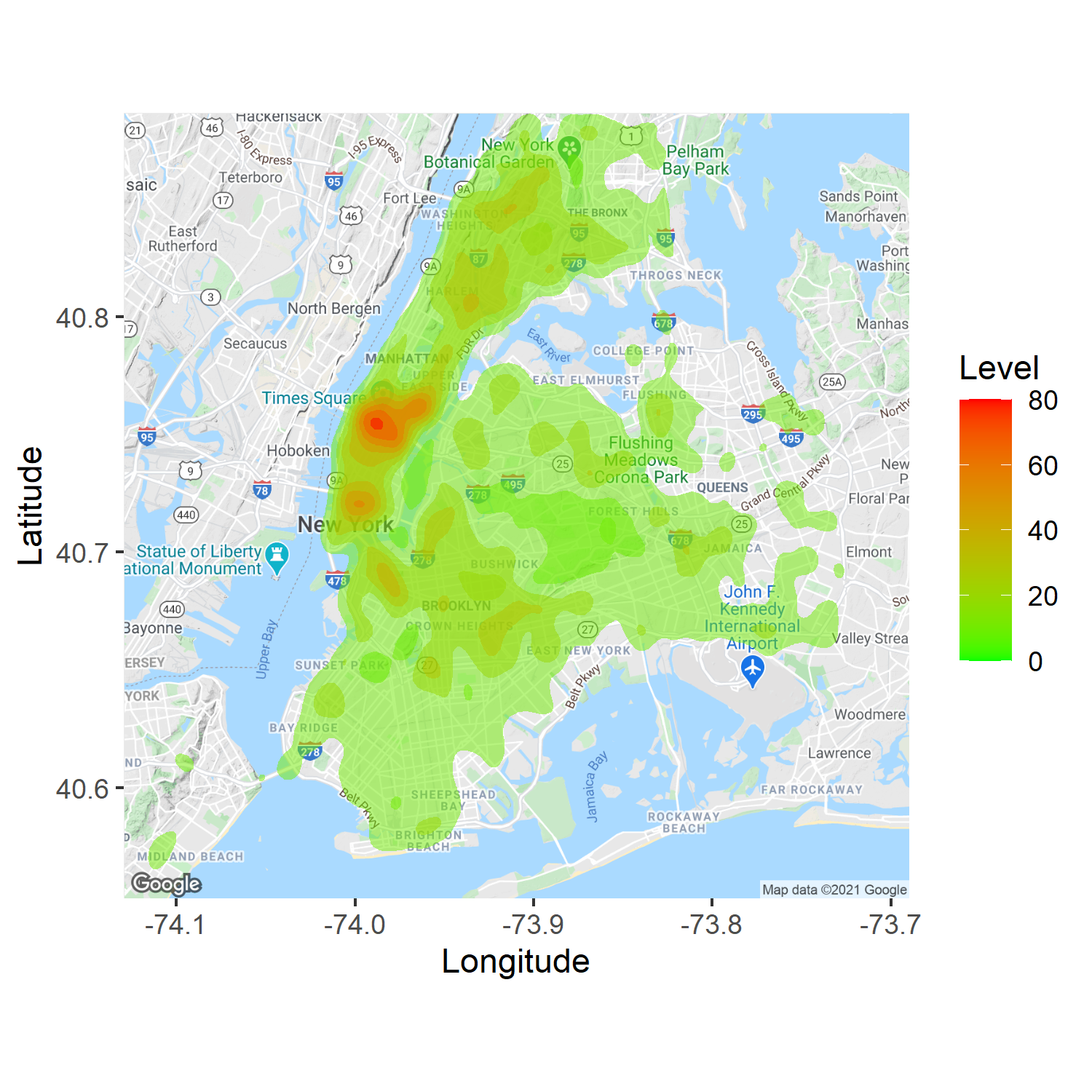}
	\end{minipage}}
 \hfill 	
  \subfloat[NYC June 2019]{
	\begin{minipage}[c][1\width]{
	   0.2\textwidth}
	   \centering
	   \includegraphics[width=1.2\textwidth]{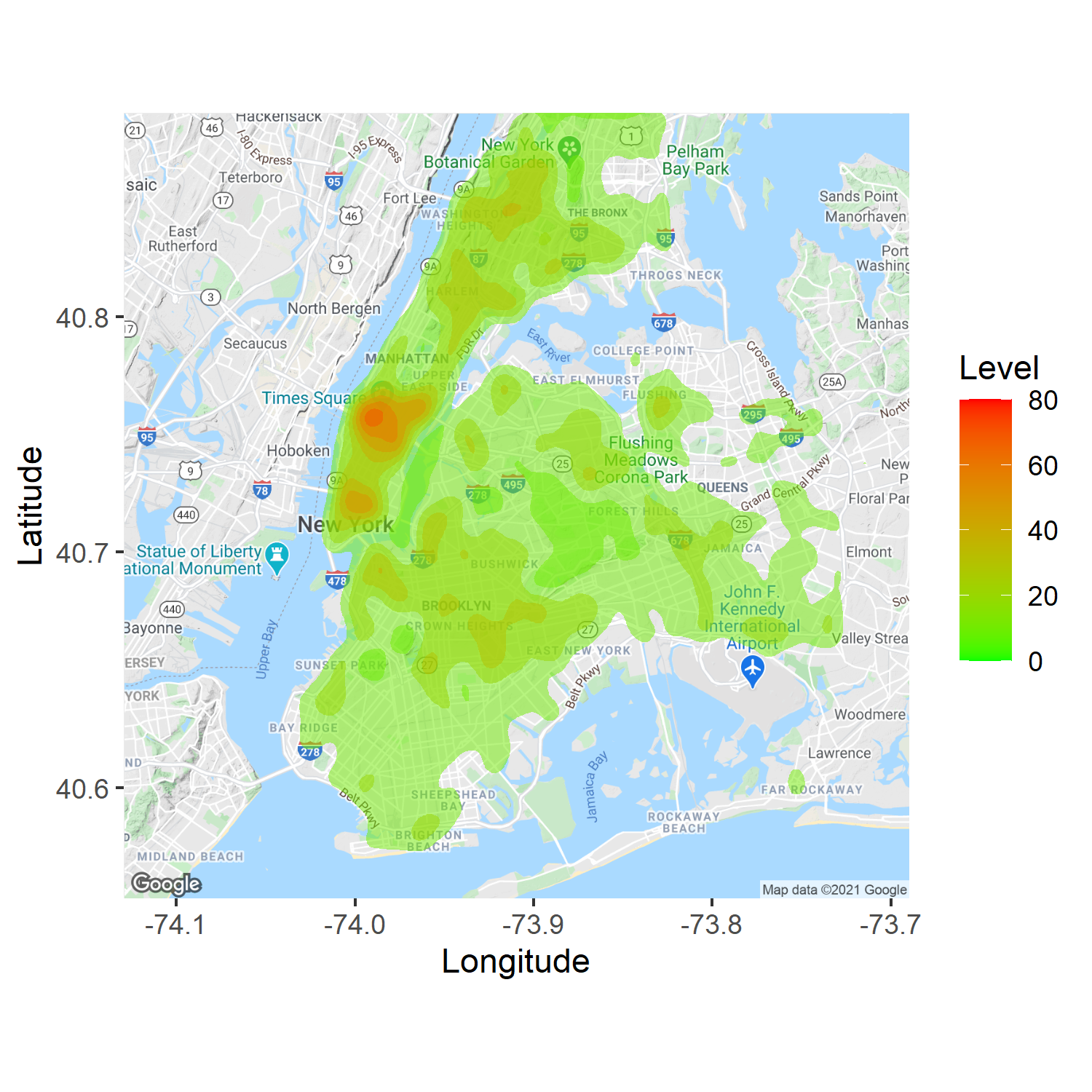}
	\end{minipage}}
 \hfill	
  \subfloat[NYC June 2020]{
	\begin{minipage}[c][1\width]{
	   0.2\textwidth}
	   \centering
	   \includegraphics[width=1.2\textwidth]{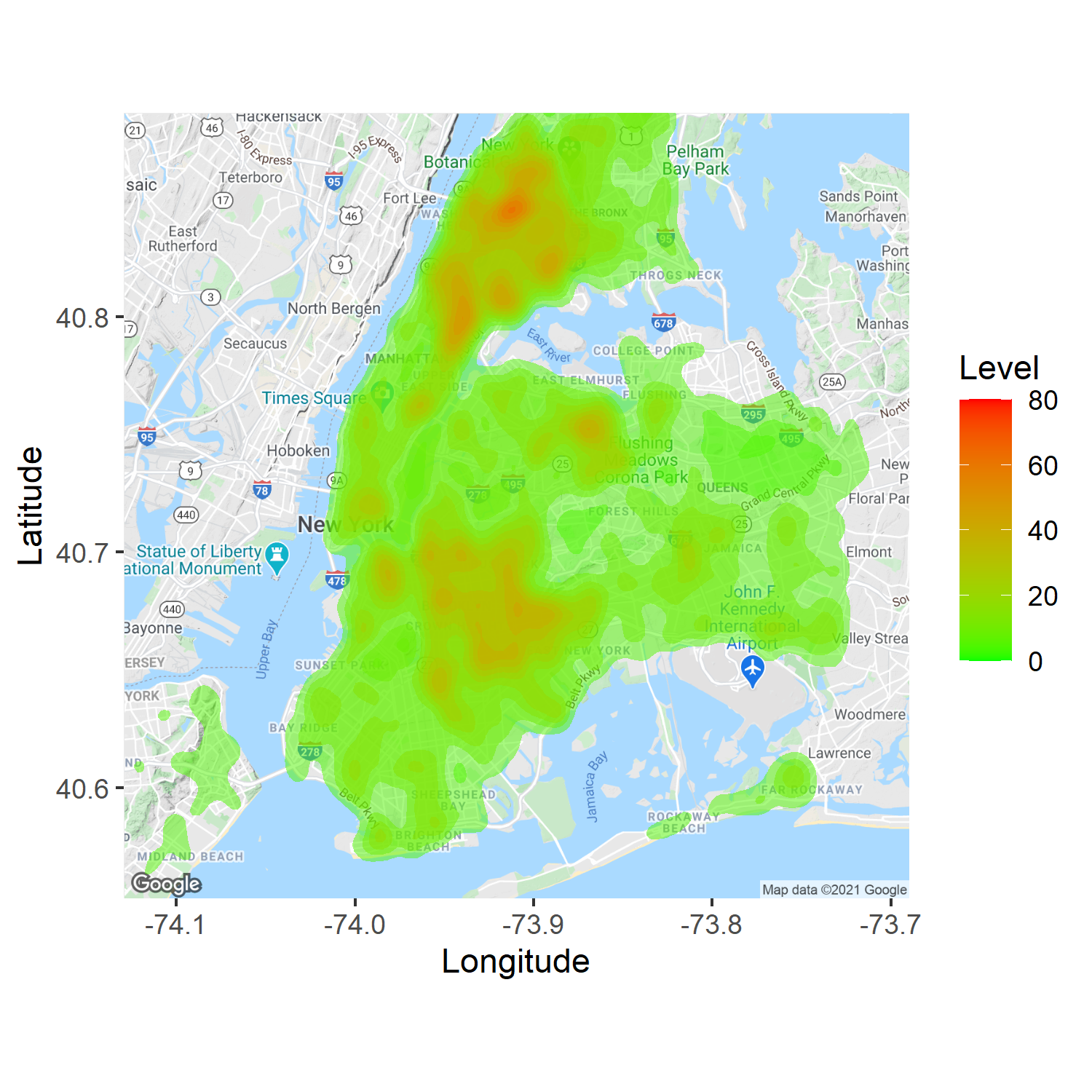}
	\end{minipage}}
 \hfill	
  \subfloat[NYC June 2021]{
	\begin{minipage}[c][1\width]{
	   0.2\textwidth}
	   \centering
	   \includegraphics[width=1.2\textwidth]{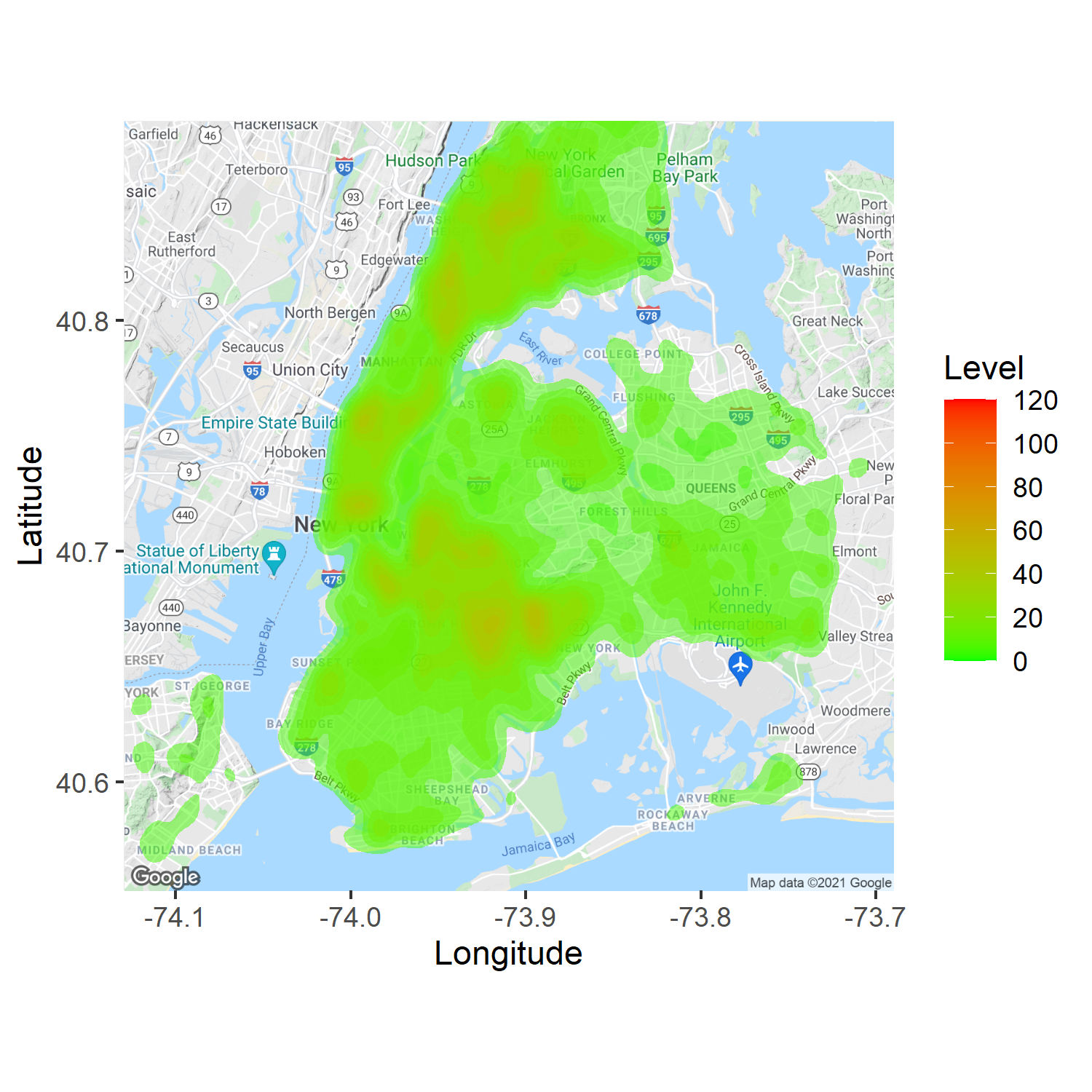}
	\end{minipage}}
\caption{The KDE results of traffic accidents in New York City. Four 30-day analysis of June, 2018--2021 are shown. The accident hotspots have shifted from Midtown and Lower Manhattan to Upper East Side, West Bronx, and southern Brooklyn.}
\end{figure}

\begin{figure}[H]
  \subfloat[LA January 2018]{
	\begin{minipage}[c][1\width]{
	   0.2\textwidth}
	   \centering
	   \includegraphics[width=1.2\textwidth]{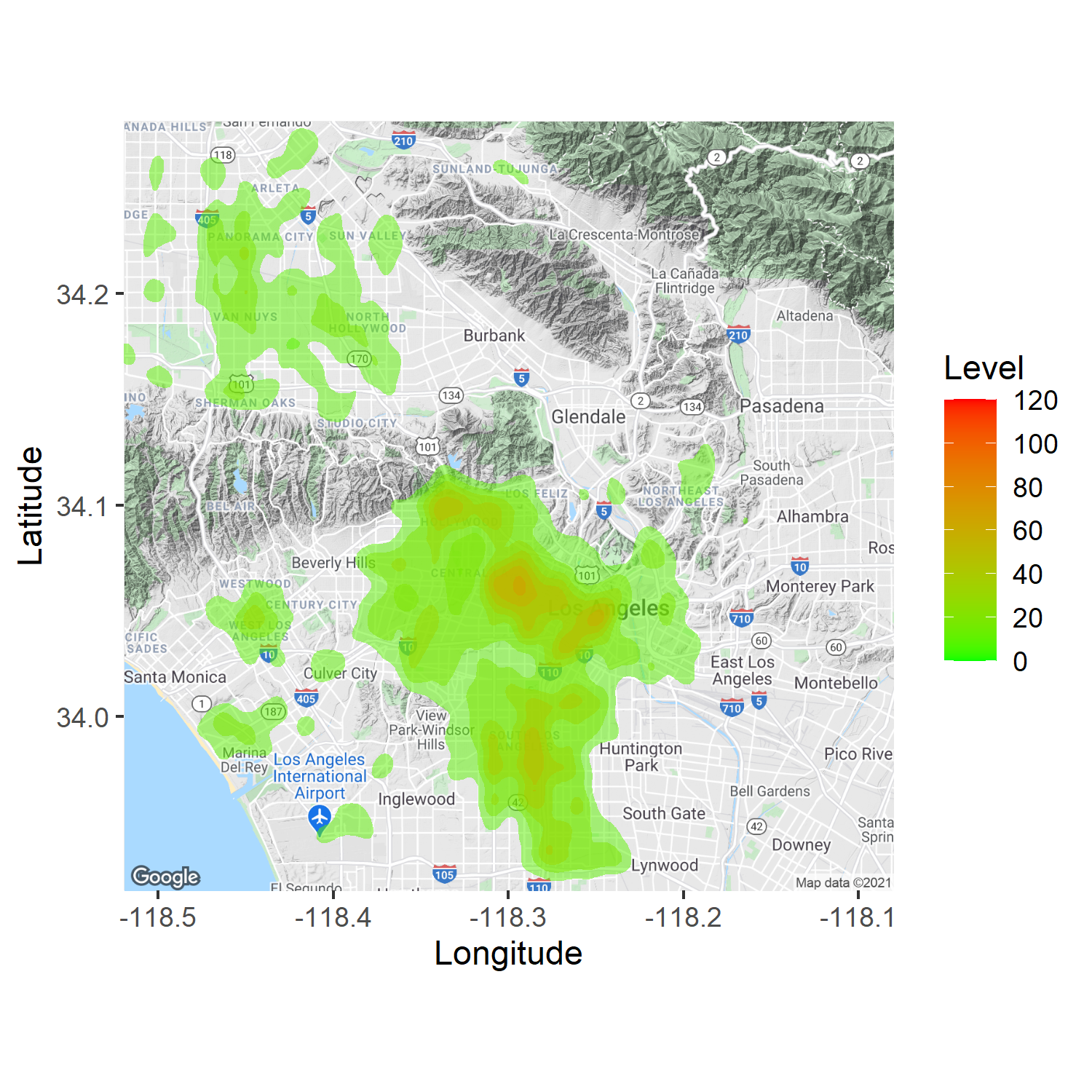}
	\end{minipage}}
 \hfill 	
  \subfloat[LA January 2019]{
	\begin{minipage}[c][1\width]{
	   0.2\textwidth}
	   \centering
	   \includegraphics[width=1.2\textwidth]{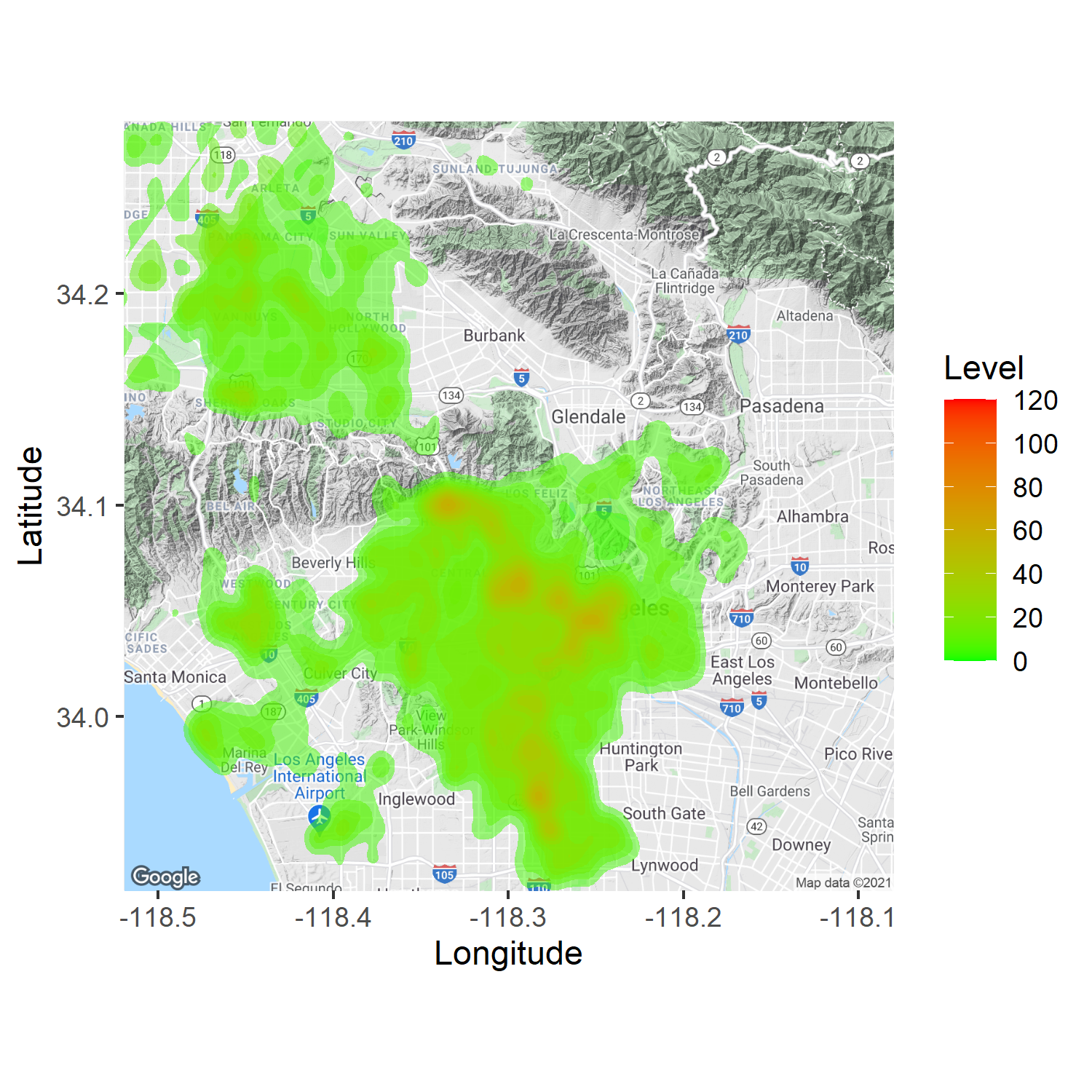}
	\end{minipage}}
 \hfill	
  \subfloat[LA January 2020]{
	\begin{minipage}[c][1\width]{
	   0.2\textwidth}
	   \centering
	   \includegraphics[width=1.2\textwidth]{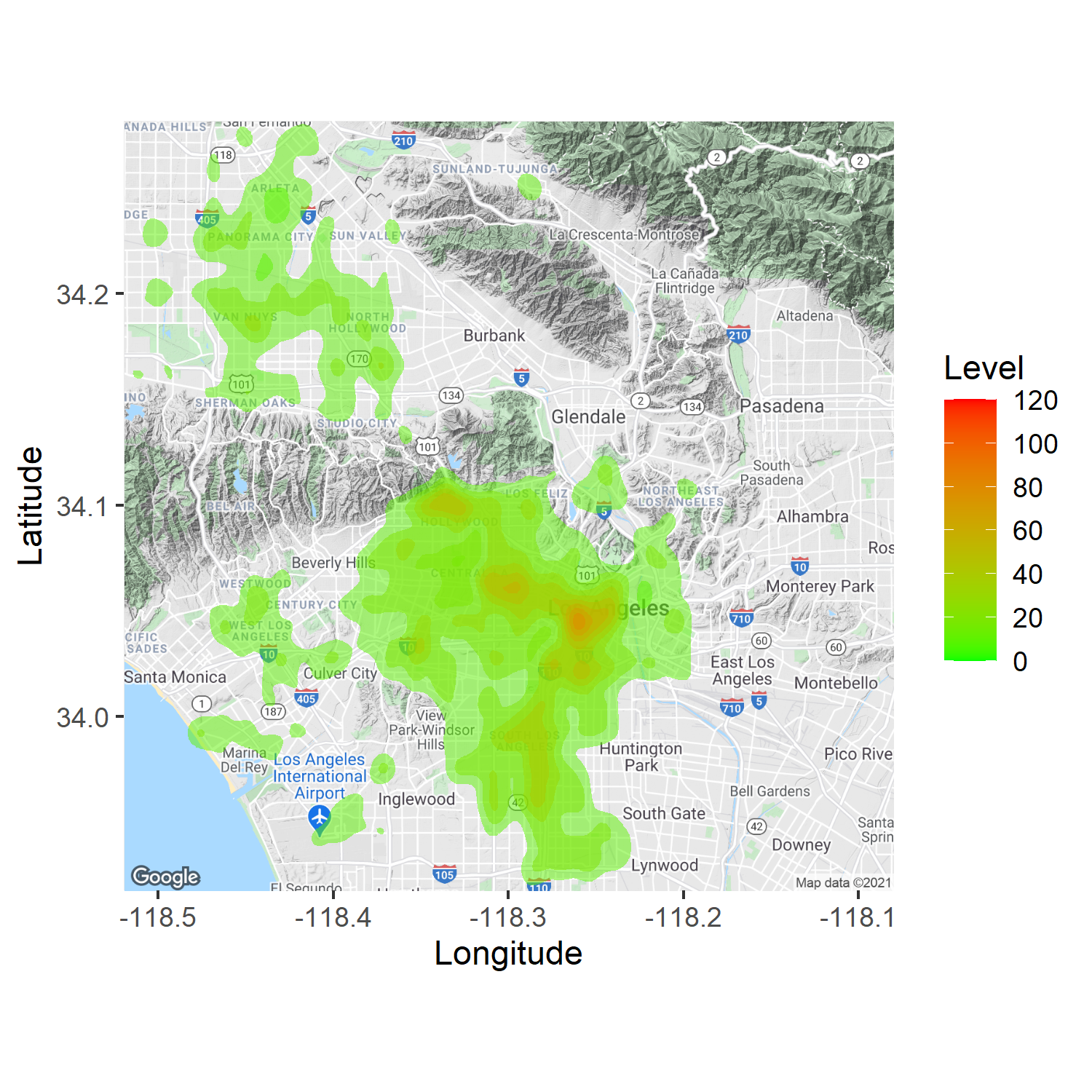}
	\end{minipage}}
 \hfill	
  \subfloat[LA January 2021]{
	\begin{minipage}[c][1\width]{
	   0.2\textwidth}
	   \centering
	   \includegraphics[width=1.2\textwidth]{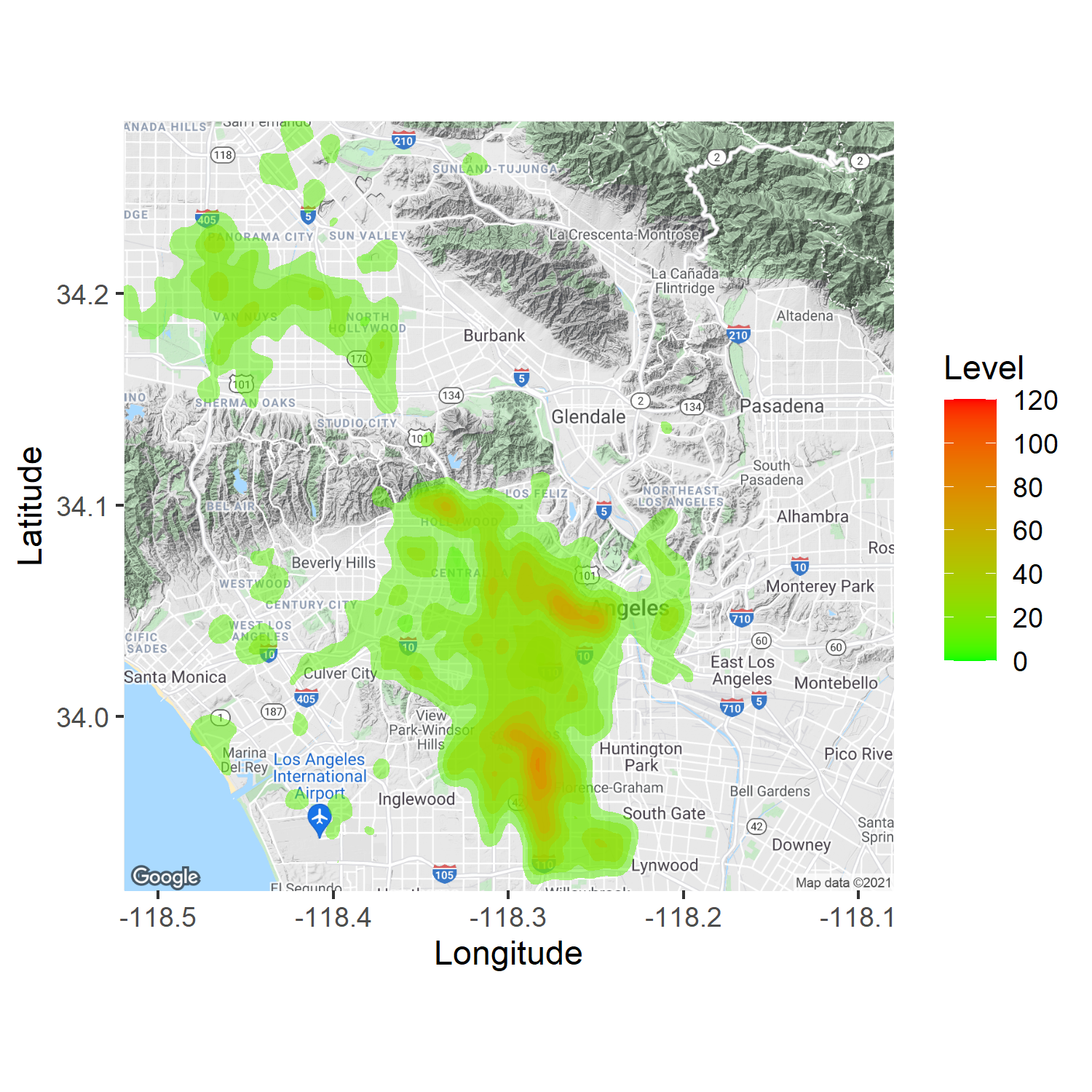}
	\end{minipage}}
\caption{The KDE results of traffic accidents in Los Angeles. Four 30-day analysis of March, 2018--2021 are shown. The accident hotspots have shifted from the Hollywood area and northern LA to southern LA.}
\end{figure}

\begin{figure}[H]
  \subfloat[LA April 2018]{
	\begin{minipage}[c][1\width]{
	   0.2\textwidth}
	   \centering
	   \includegraphics[width=1.2\textwidth]{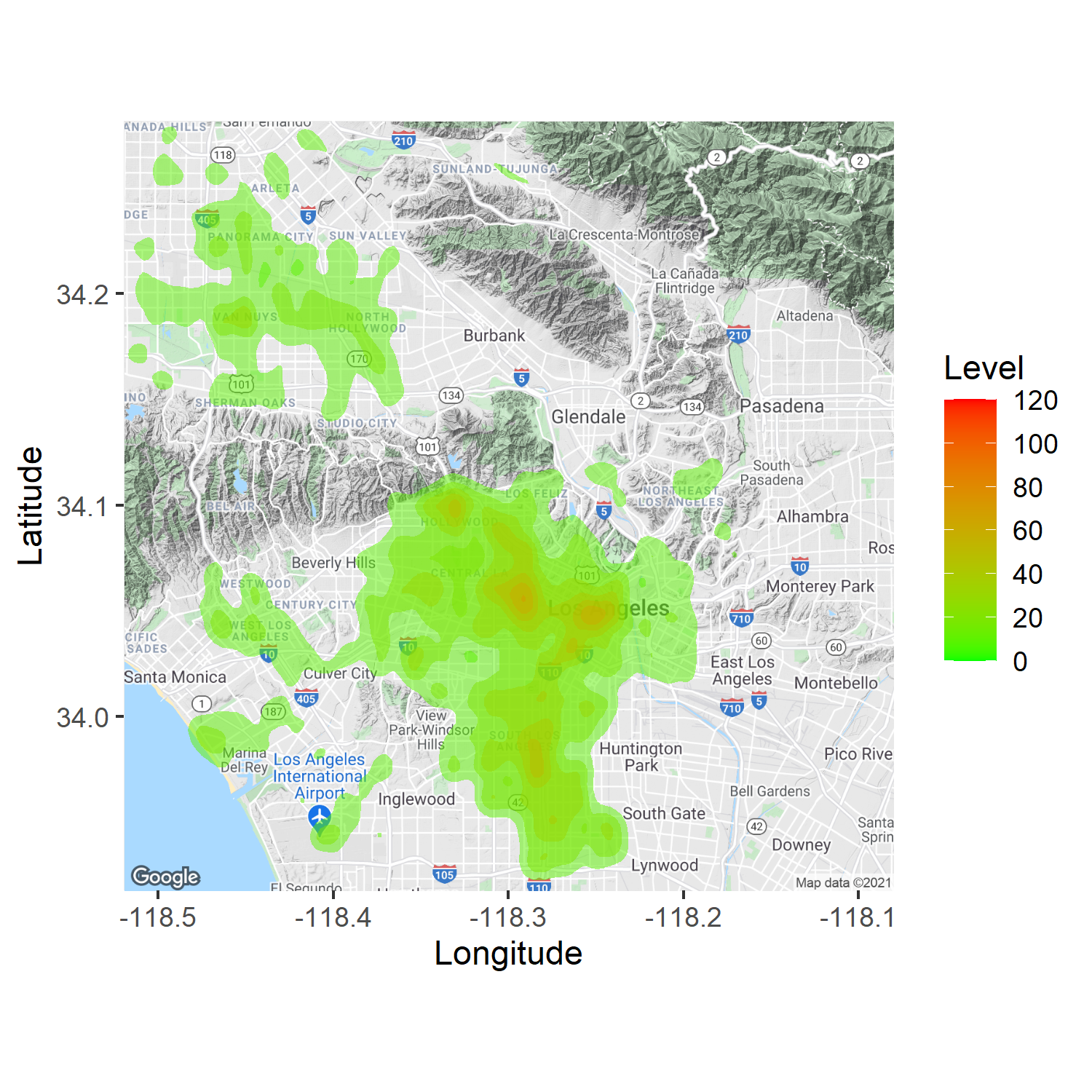}
	\end{minipage}}
 \hfill 	
  \subfloat[LA April 2019]{
	\begin{minipage}[c][1\width]{
	   0.2\textwidth}
	   \centering
	   \includegraphics[width=1.2\textwidth]{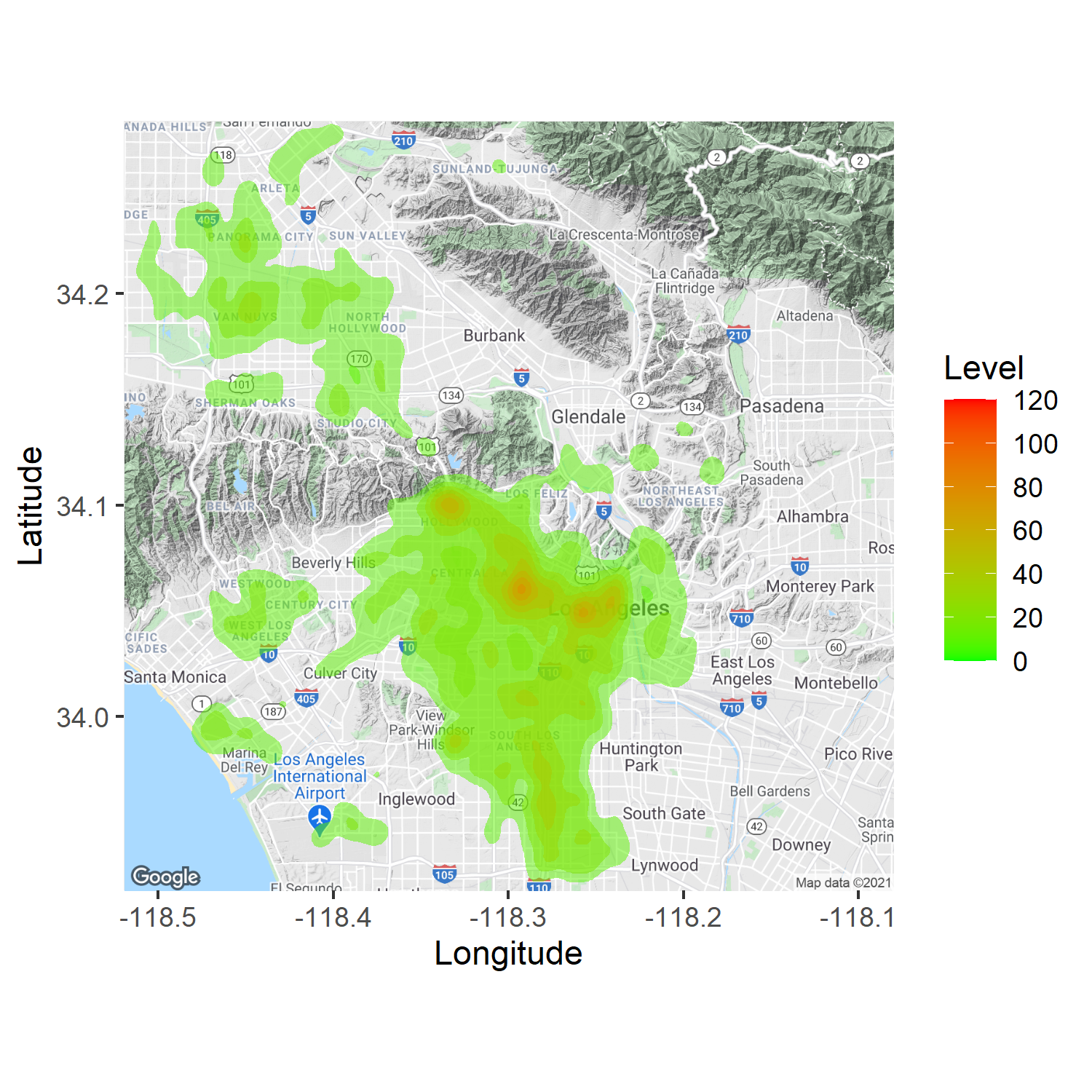}
	\end{minipage}}
 \hfill	
  \subfloat[LA April 2020]{
	\begin{minipage}[c][1\width]{
	   0.2\textwidth}
	   \centering
	   \includegraphics[width=1.2\textwidth]{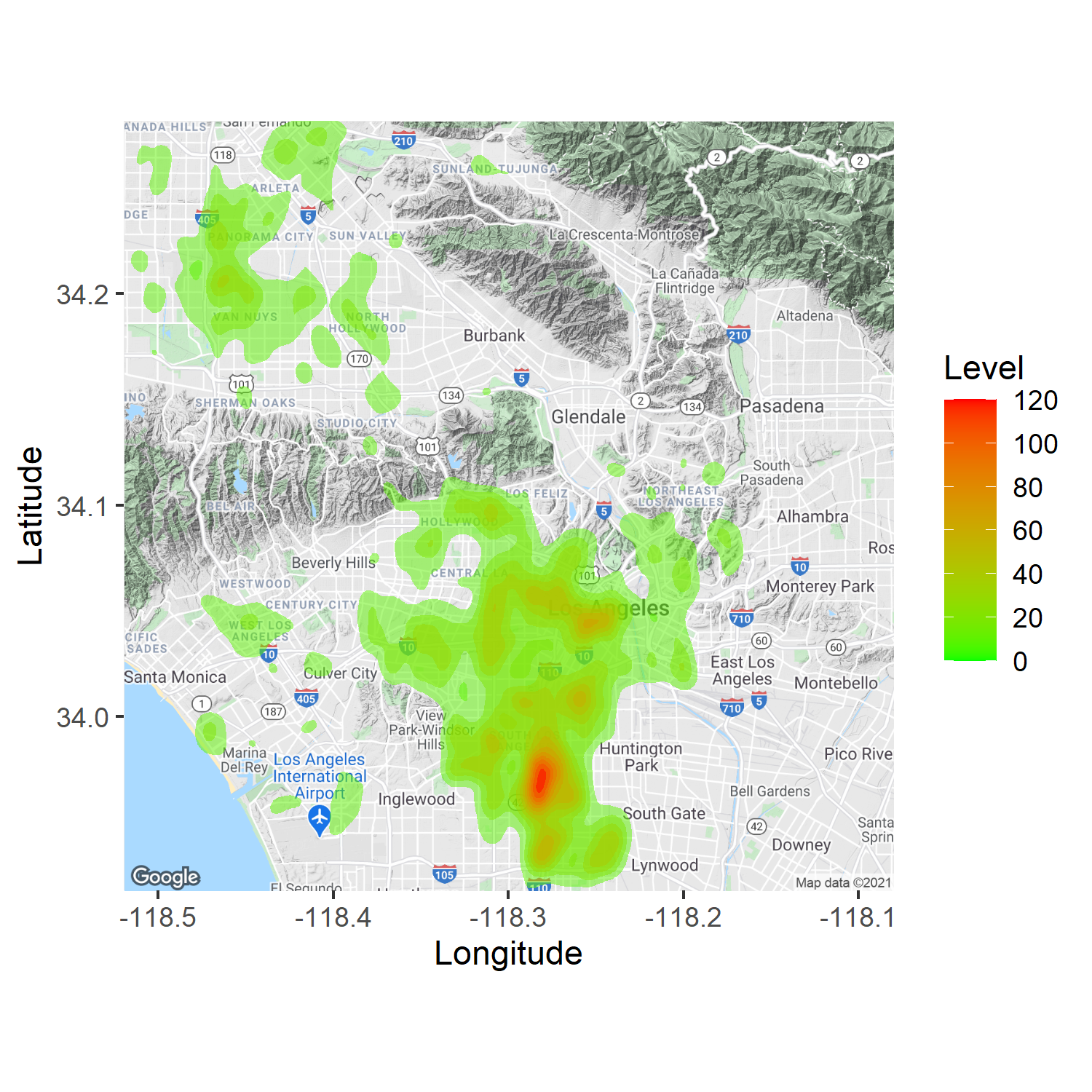}
	\end{minipage}}
 \hfill	
  \subfloat[LA April 2021]{
	\begin{minipage}[c][1\width]{
	   0.2\textwidth}
	   \centering
	   \includegraphics[width=1.2\textwidth]{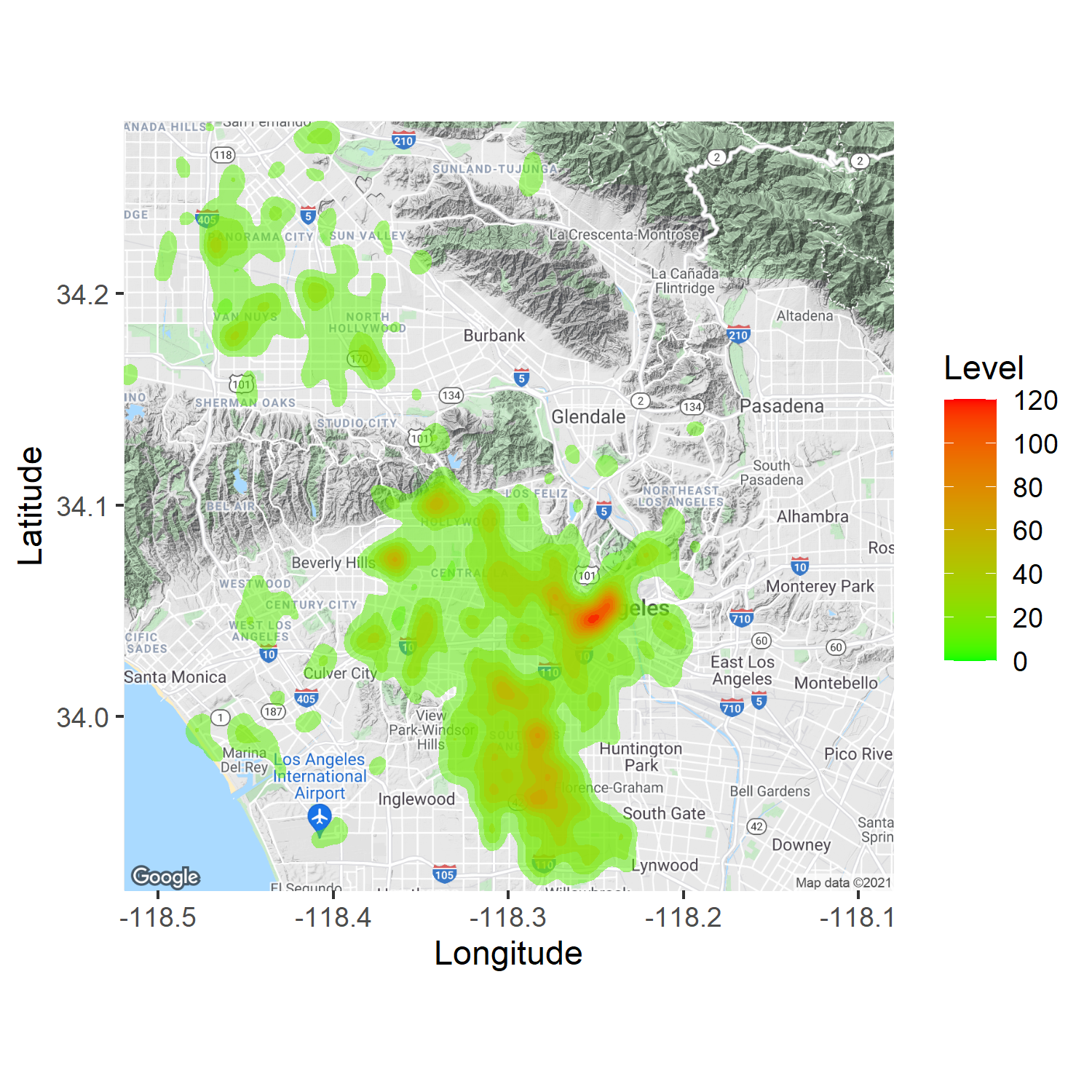}
	\end{minipage}}
\caption{The KDE results of traffic accidents in Los Angeles. Four 30-day analysis of April, 2018--2021 are shown. The accident hotspots have shifted from the Hollywood area and northern LA to southern LA. The hotspots have become more prominent in southern LA in 2020 and towards both southern and northern LA in 2021.} 
\end{figure}

\begin{figure}[H]
  \subfloat[LA June 2018]{
	\begin{minipage}[c][1\width]{
	   0.2\textwidth}
	   \centering
	   \includegraphics[width=1.2\textwidth]{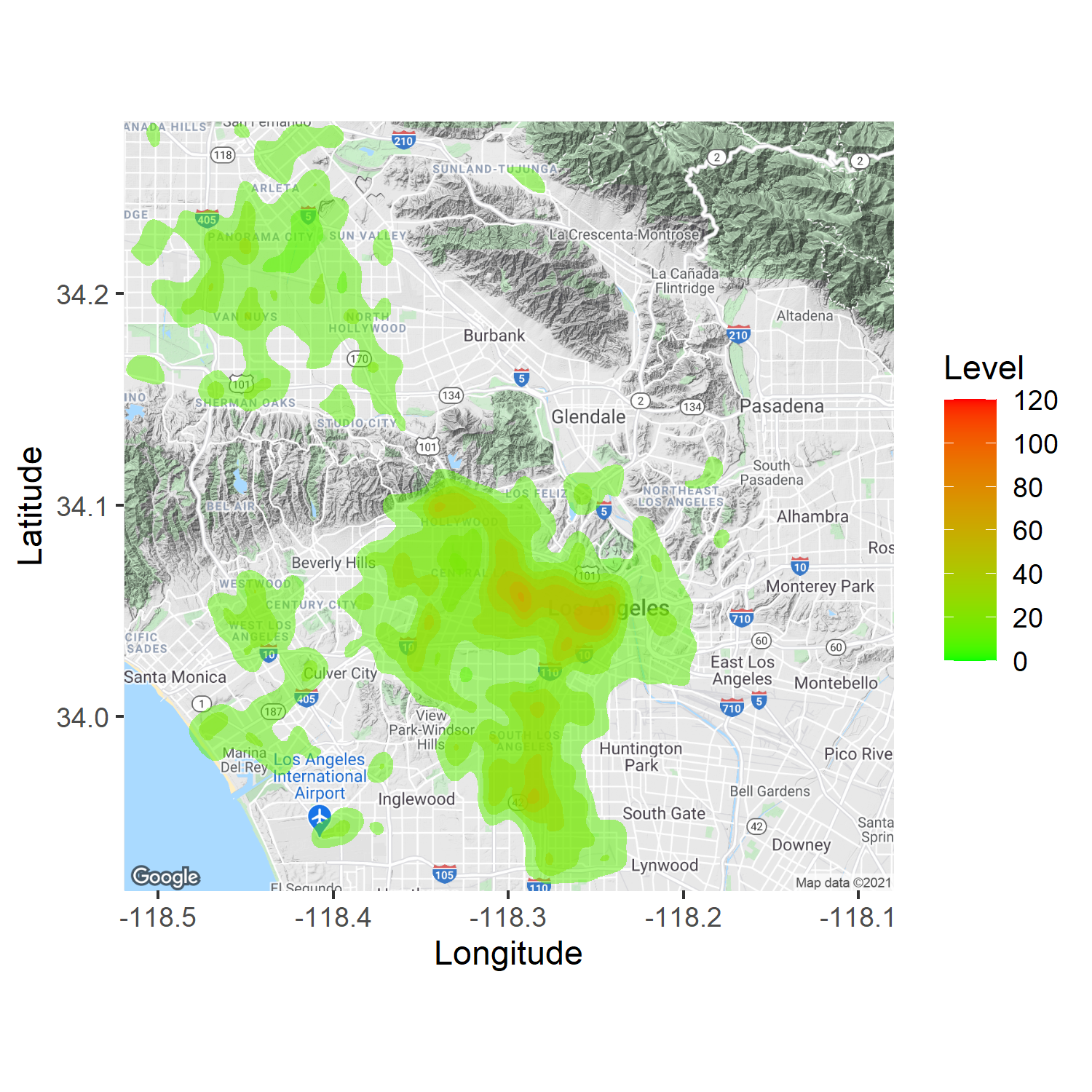}
	\end{minipage}}
 \hfill 	
  \subfloat[LA June 2019]{
	\begin{minipage}[c][1\width]{
	   0.2\textwidth}
	   \centering
	   \includegraphics[width=1.2\textwidth]{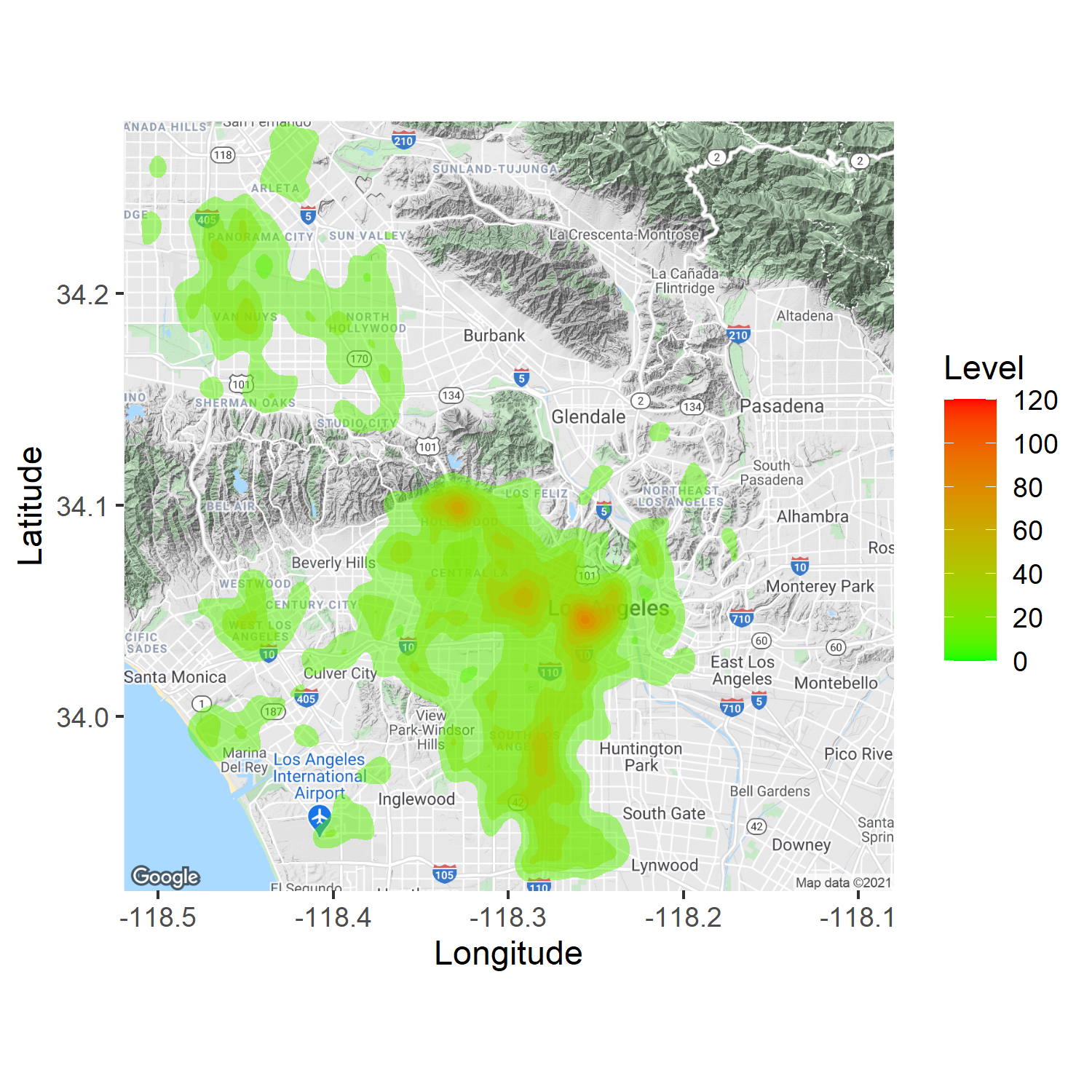}
	\end{minipage}}
 \hfill	
  \subfloat[LA June 2020]{
	\begin{minipage}[c][1\width]{
	   0.2\textwidth}
	   \centering
	   \includegraphics[width=1.2\textwidth]{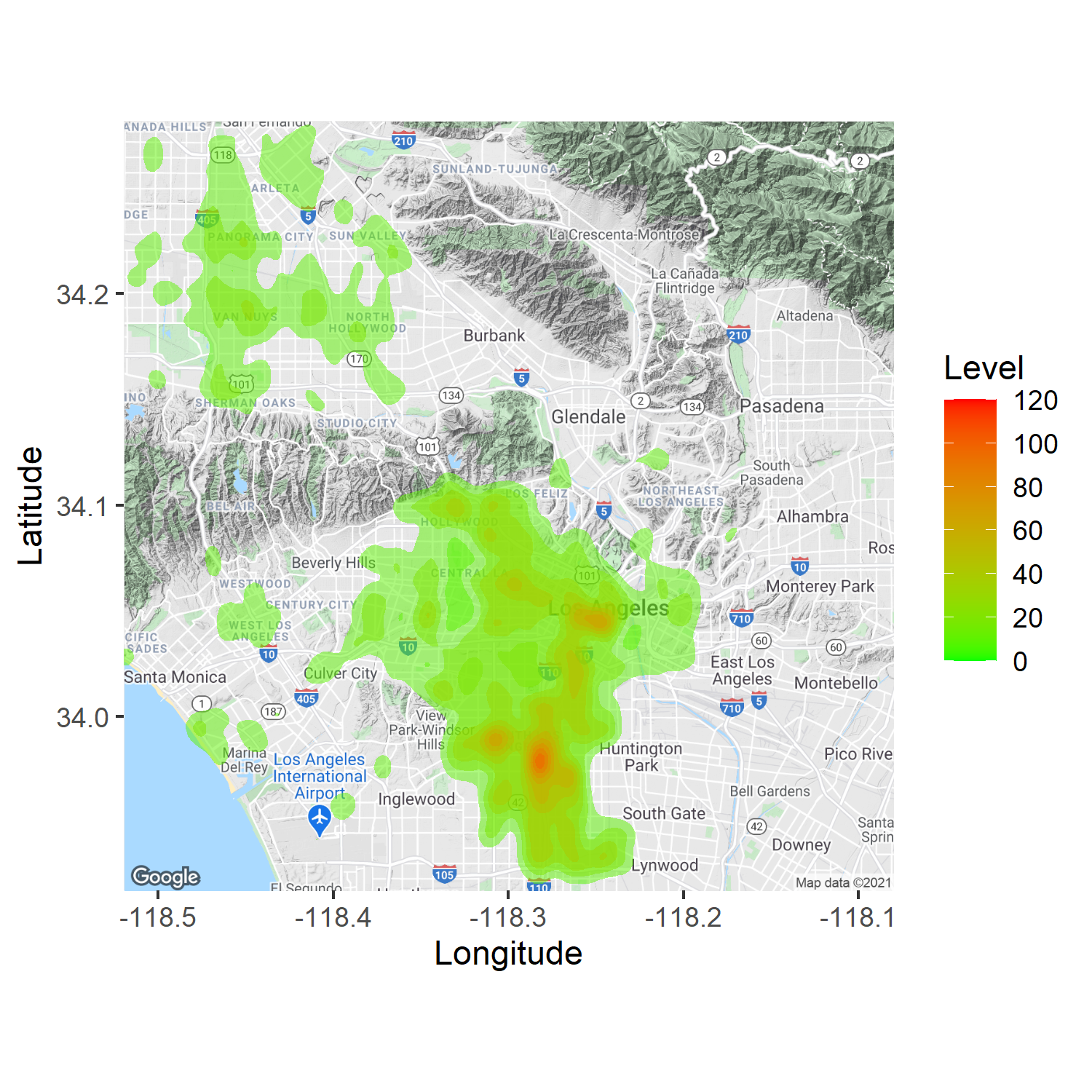}
	\end{minipage}}
 \hfill	
  \subfloat[LA June 2021]{
	\begin{minipage}[c][1\width]{
	   0.2\textwidth}
	   \centering
	   \includegraphics[width=1.2\textwidth]{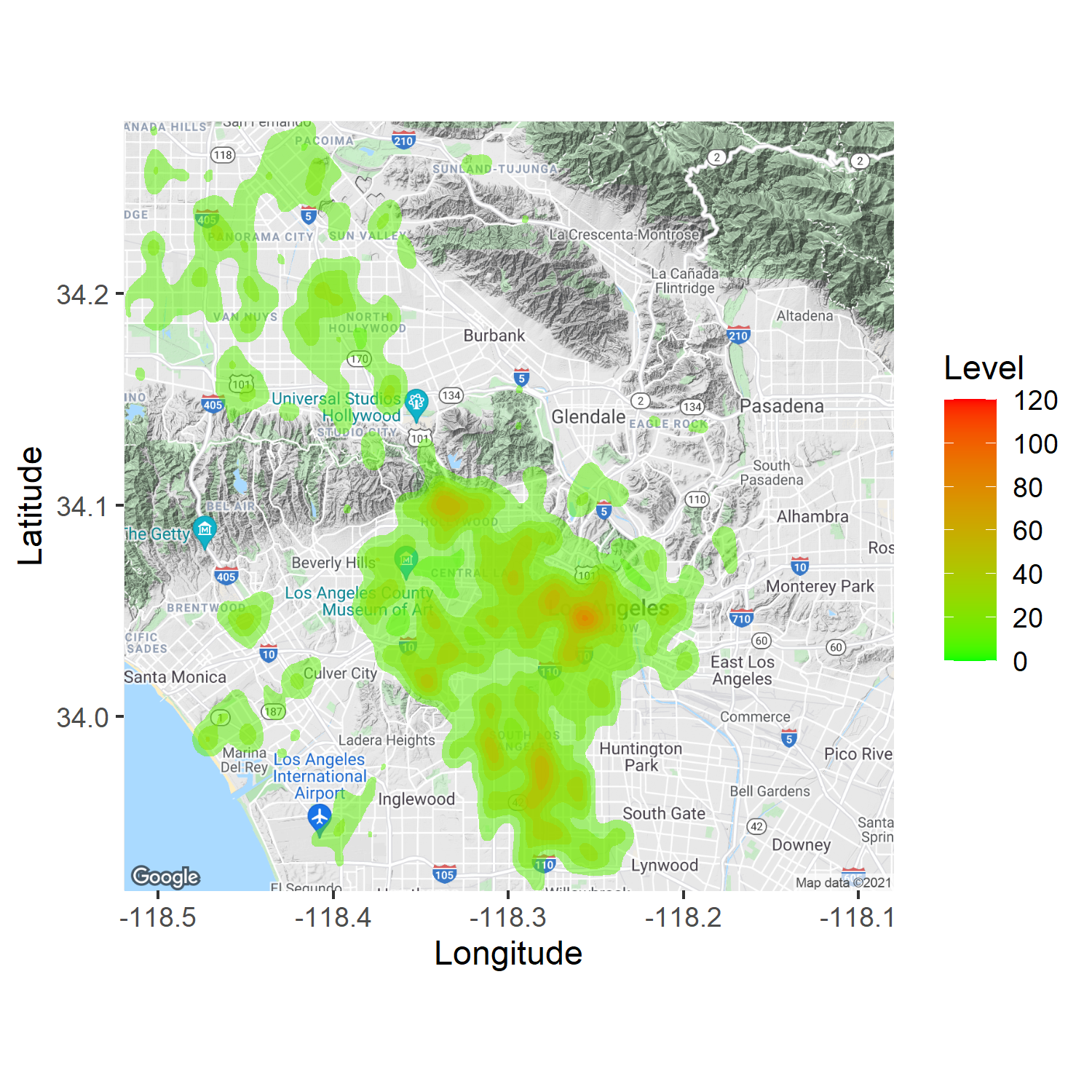}
	\end{minipage}}
\caption{The KDE results of traffic accidents in Los Angeles. Four 30-day analysis of June, 2018--2021 are shown. The accident hotspots have shifted from the Hollywood area and northern LA to southern LA. The hotspots are shifted towards the southern LA from northern LA. }
\end{figure}

\begin{figure}[H]
  \subfloat[BOSTON January 2018]{
	\begin{minipage}[c][1\width]{
	   0.2\textwidth}
	   \centering
	   \includegraphics[width=1.2\textwidth]{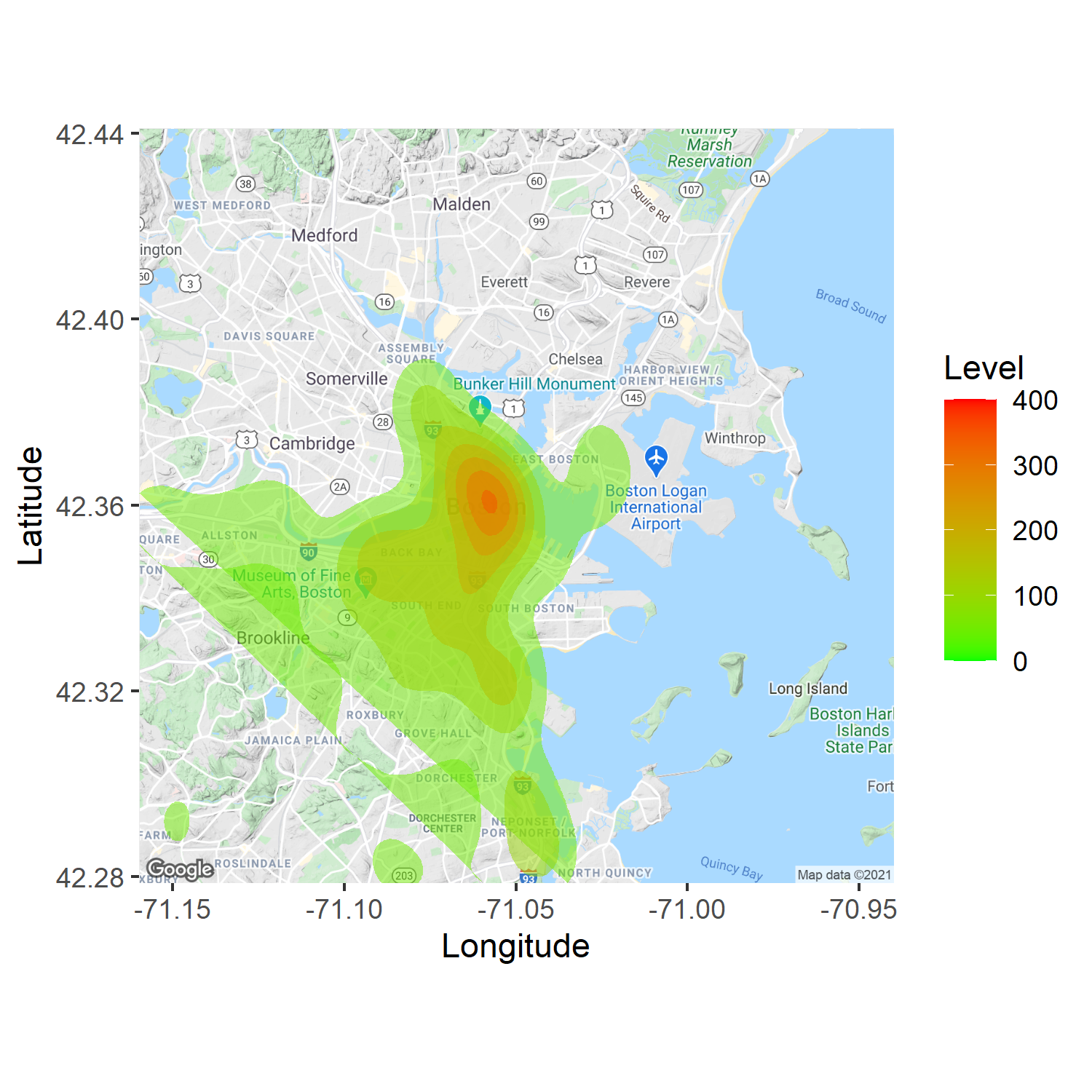}
	\end{minipage}}
 \hfill 	
  \subfloat[BOSTON January 2019]{
	\begin{minipage}[c][1\width]{
	   0.2\textwidth}
	   \centering
	   \includegraphics[width=1.2\textwidth]{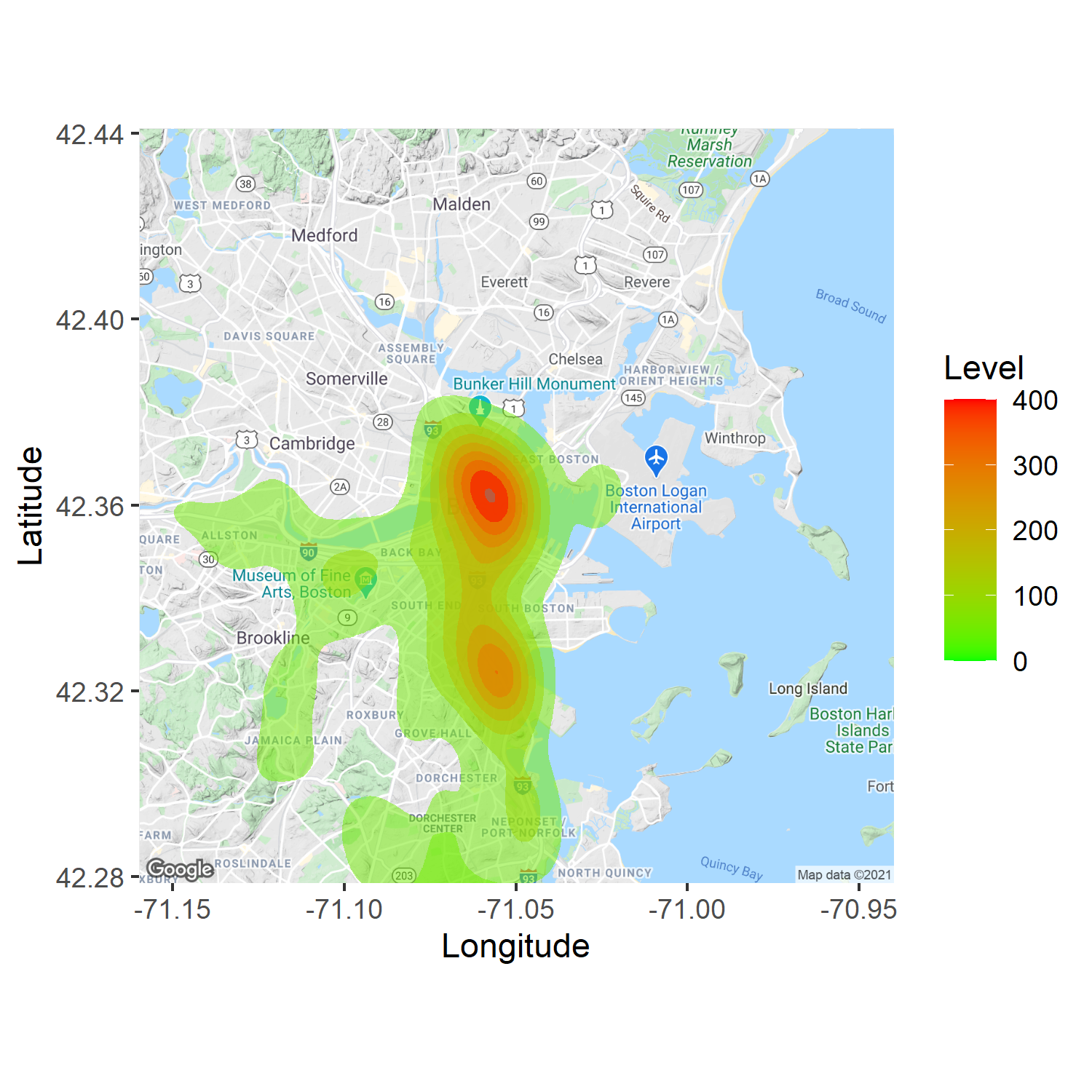}
	\end{minipage}}
 \hfill	
  \subfloat[BOSTON January 2020]{
	\begin{minipage}[c][1\width]{
	   0.2\textwidth}
	   \centering
	   \includegraphics[width=1.2\textwidth]{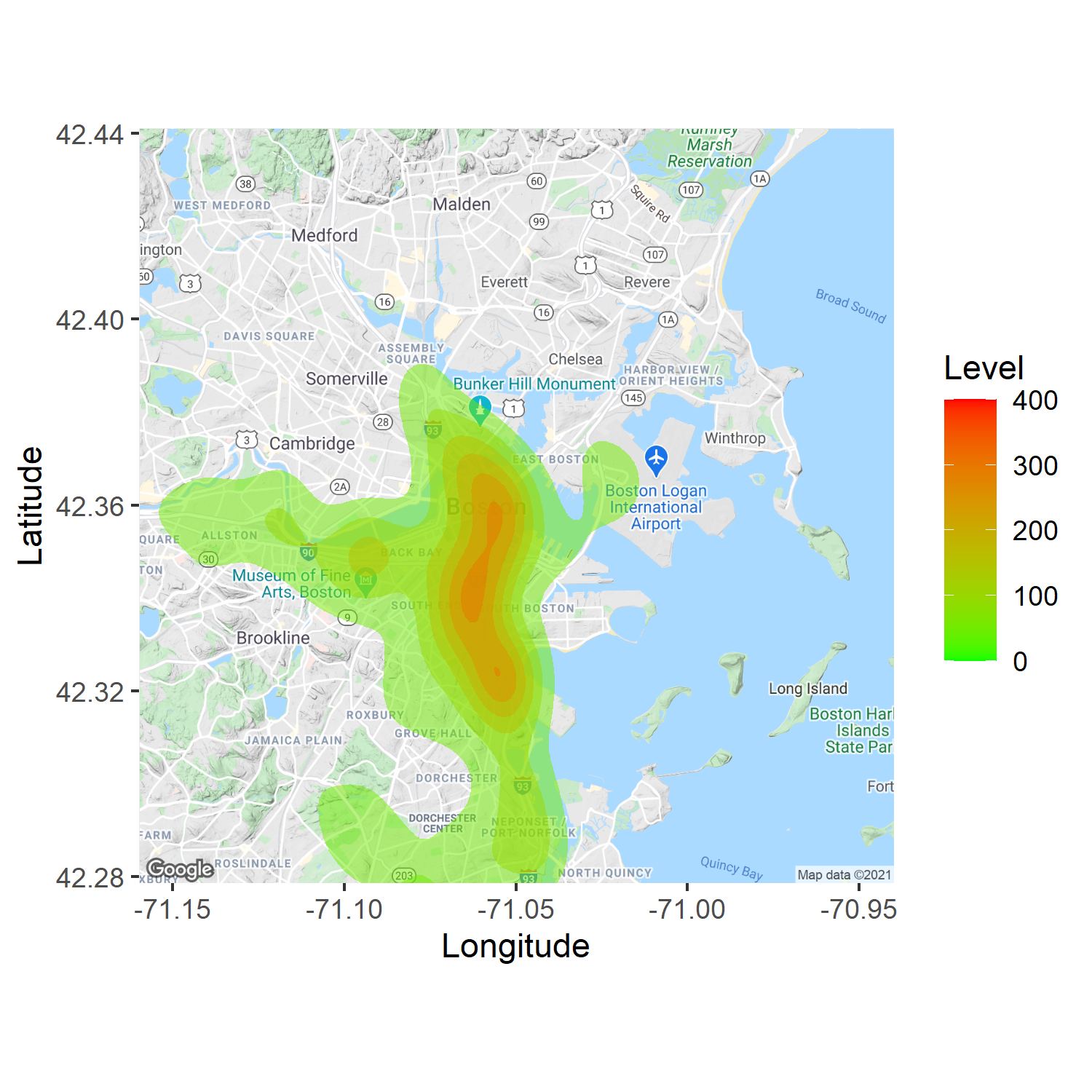}
	\end{minipage}}
 \hfill	
  \subfloat[BOSTON January 2021]{
	\begin{minipage}[c][1\width]{
	   0.2\textwidth}
	   \centering
	   \includegraphics[width=1.2\textwidth]{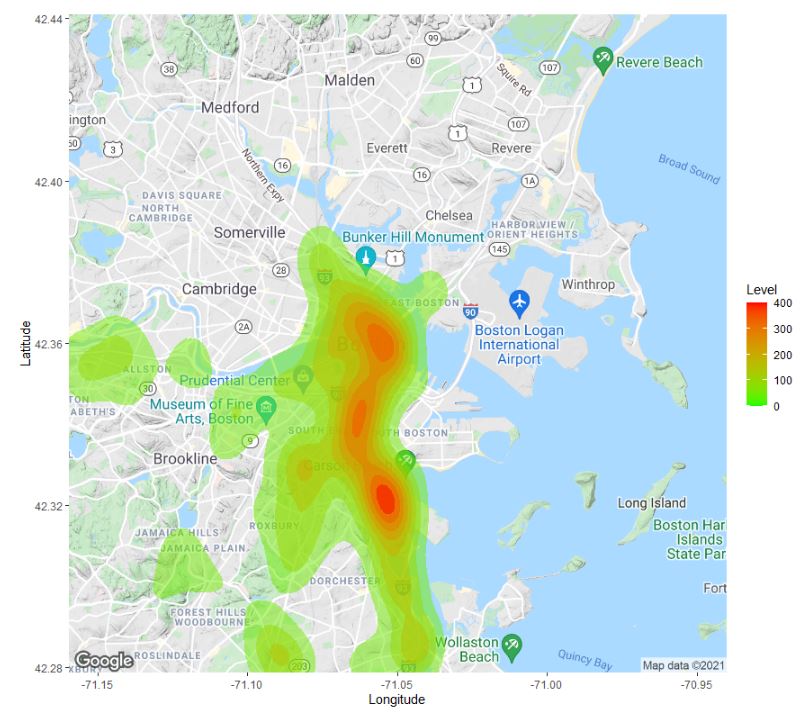}
	\end{minipage}}
\caption{The KDE results of traffic accidents in Boston. Four 30-day analysis of January, 2018--2021 are shown. The accident hotspots have shifted from South End and South Boston to North and South Dorchester.}
\end{figure}

\begin{figure}[H]
  \subfloat[BOSTON June 2018]{
	\begin{minipage}[c][1\width]{
	   0.2\textwidth}
	   \centering
	   \includegraphics[width=1.2\textwidth]{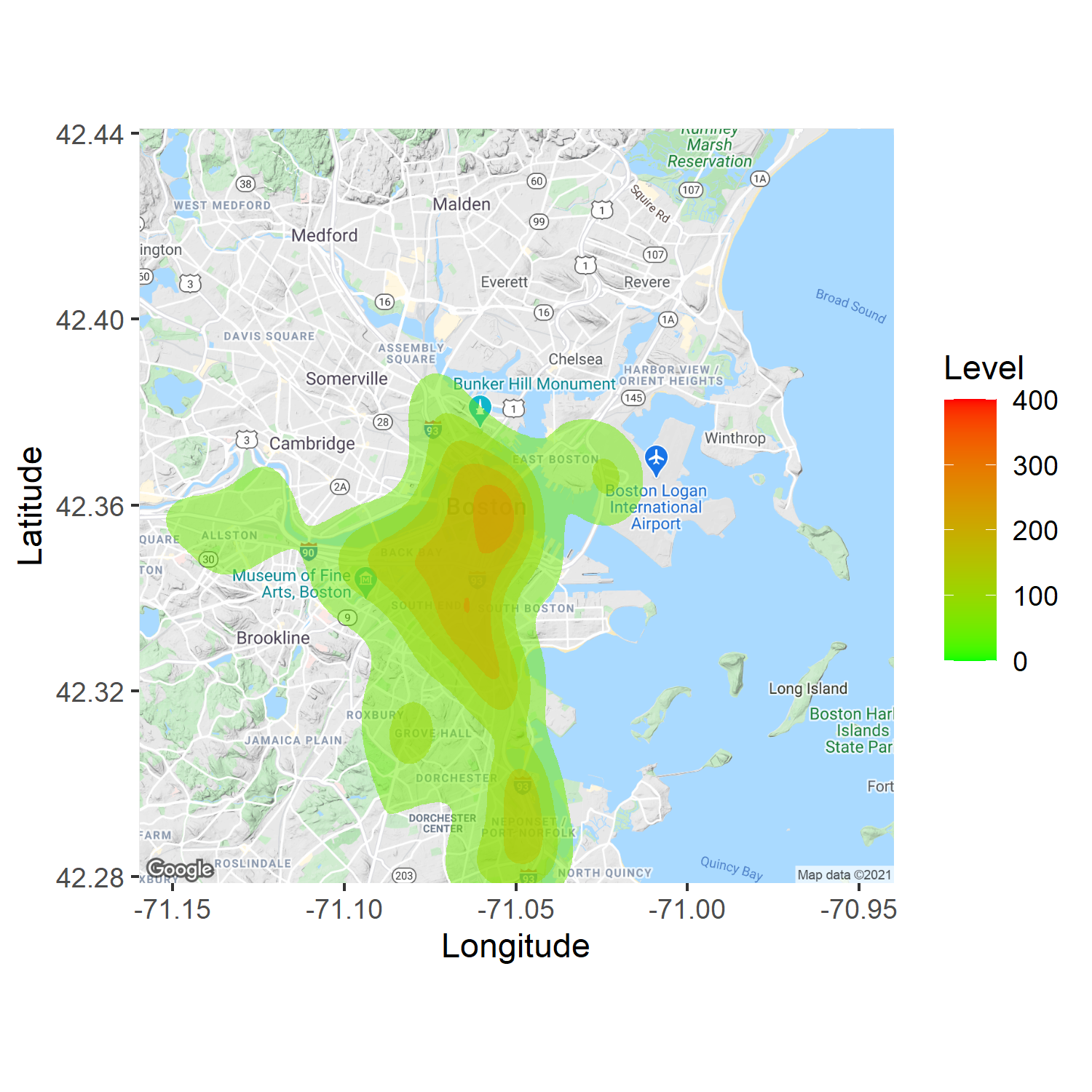}
	\end{minipage}}
 \hfill 	
  \subfloat[BOSTON June 2019]{
	\begin{minipage}[c][1\width]{
	   0.2\textwidth}
	   \centering
	   \includegraphics[width=1.2\textwidth]{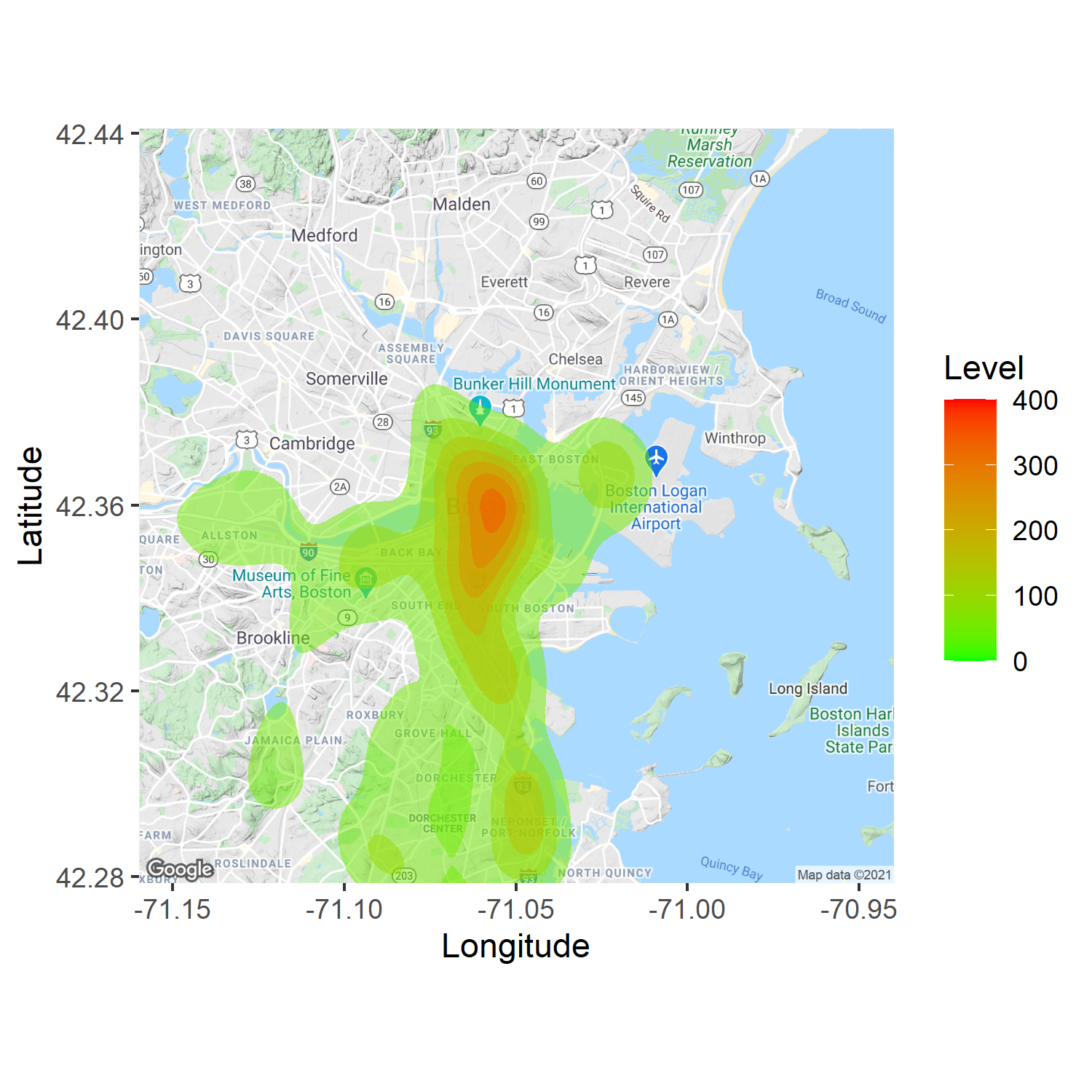}
	\end{minipage}}
 \hfill	
  \subfloat[BOSTON June 2020]{
	\begin{minipage}[c][1\width]{
	   0.2\textwidth}
	   \centering
	   \includegraphics[width=1.2\textwidth]{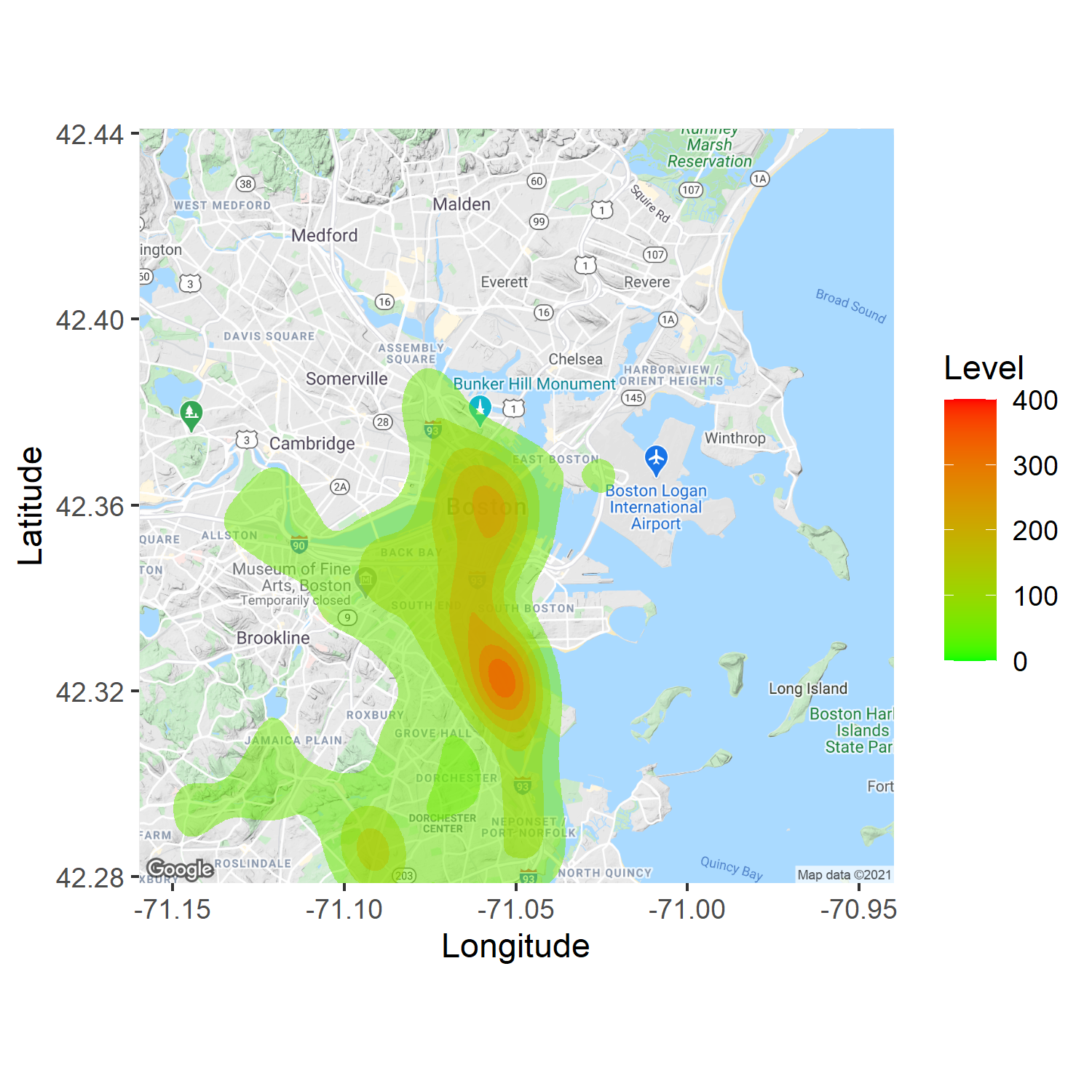}
	\end{minipage}}
 \hfill	
  \subfloat[BOSTON June 2021]{
	\begin{minipage}[c][1\width]{
	   0.2\textwidth}
	   \centering
	   \includegraphics[width=1.2\textwidth]{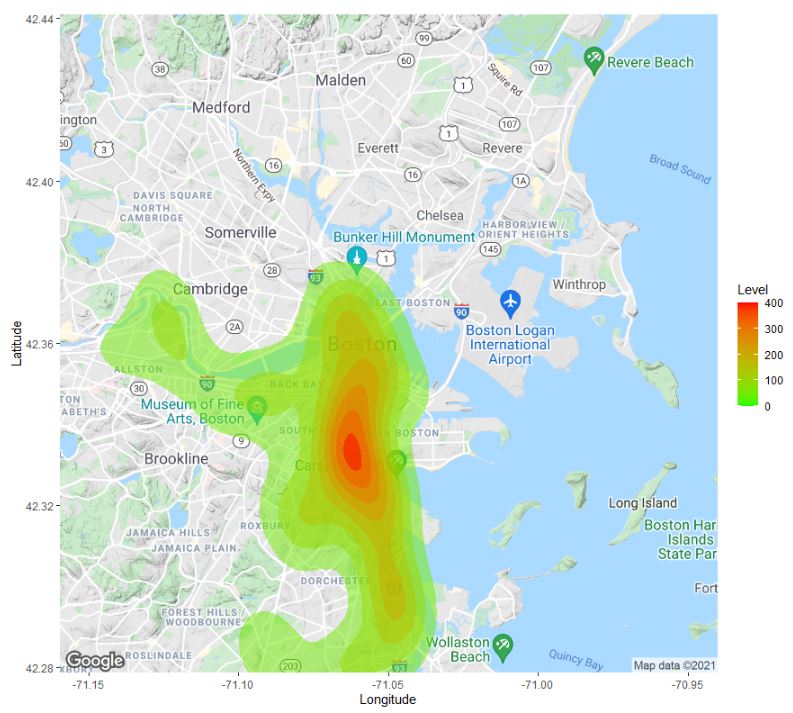}
	\end{minipage}}
\caption{The KDE results of traffic accidents in Boston. Four 30-day analysis of June, 2018--2021 are shown. The accident hotspots have shifted from South End and South Boston to North and South Dorchester.}
\end{figure}

\begin{figure}[H]
  \subfloat[Road Network in detail]{
	\begin{minipage}[c][1\width]{
	   0.2\textwidth}
	   \centering
	   \includegraphics[width=1.2\textwidth]{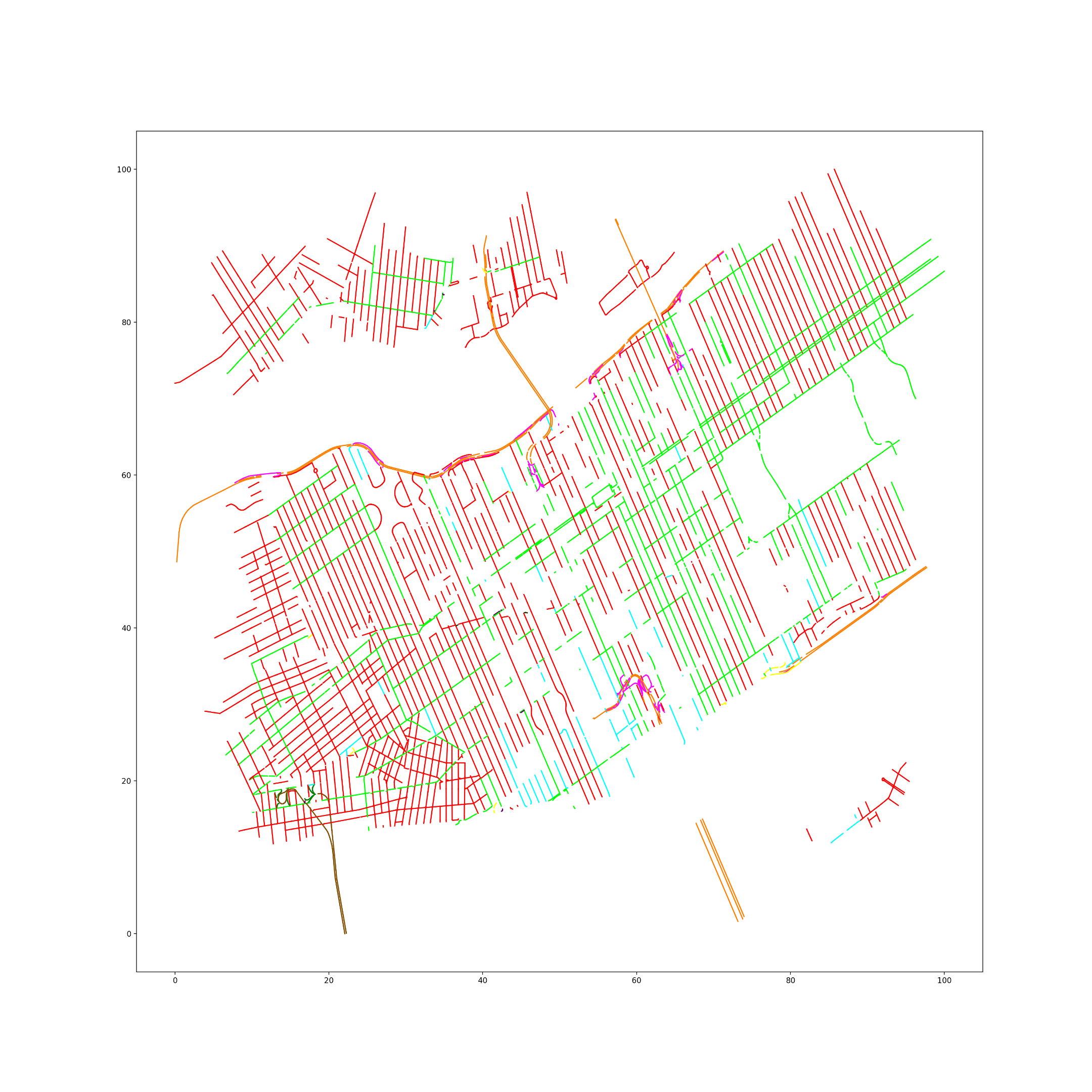}
	\end{minipage}}
 \hfill 	
  \subfloat[Type of highway percentages]{
	\begin{minipage}[c][1\width]{
	   0.2\textwidth}
	   \centering
	   \includegraphics[width=1.2\textwidth]{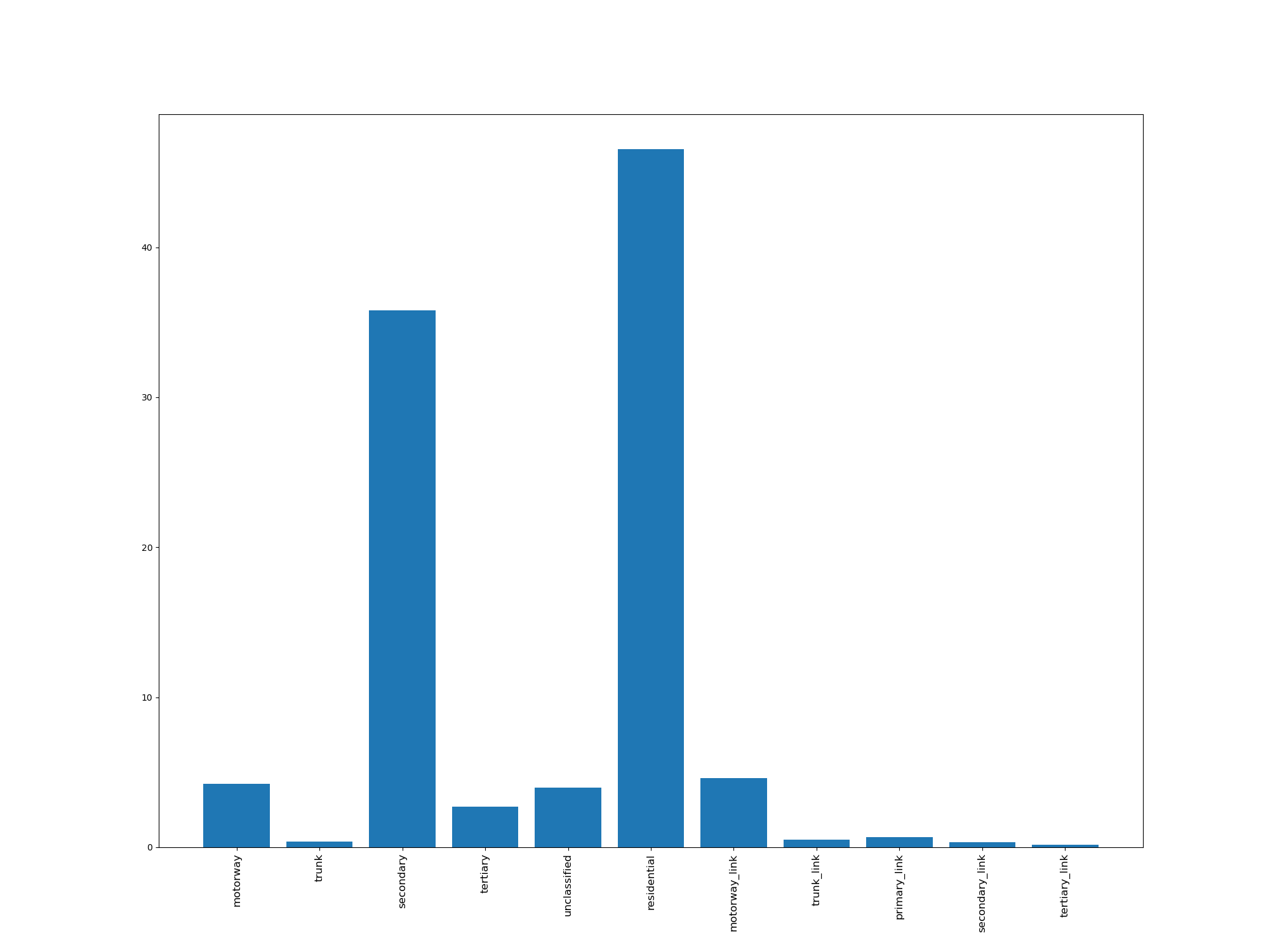}
	\end{minipage}}
 \hfill	
  \subfloat[Type of highway percentage normalized]{
	\begin{minipage}[c][1\width]{
	   0.2\textwidth}
	   \centering
	   \includegraphics[width=1.2\textwidth]{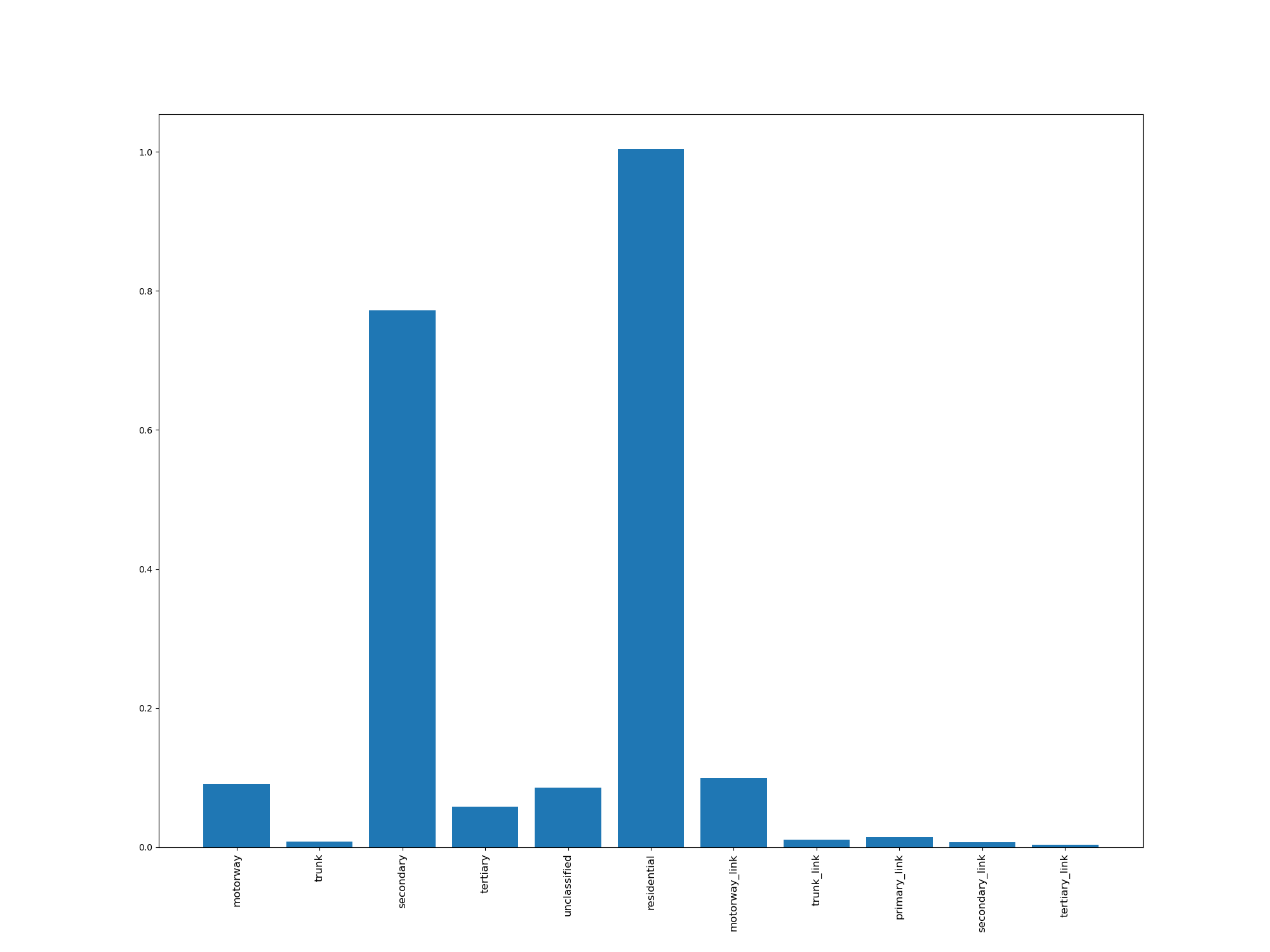}
	\end{minipage}}
 \hfill
  \subfloat[Type of Highways total distance]{
	\begin{minipage}[c][1\width]{
	   0.2\textwidth}
	   \centering
	   \includegraphics[width=1.2\textwidth]{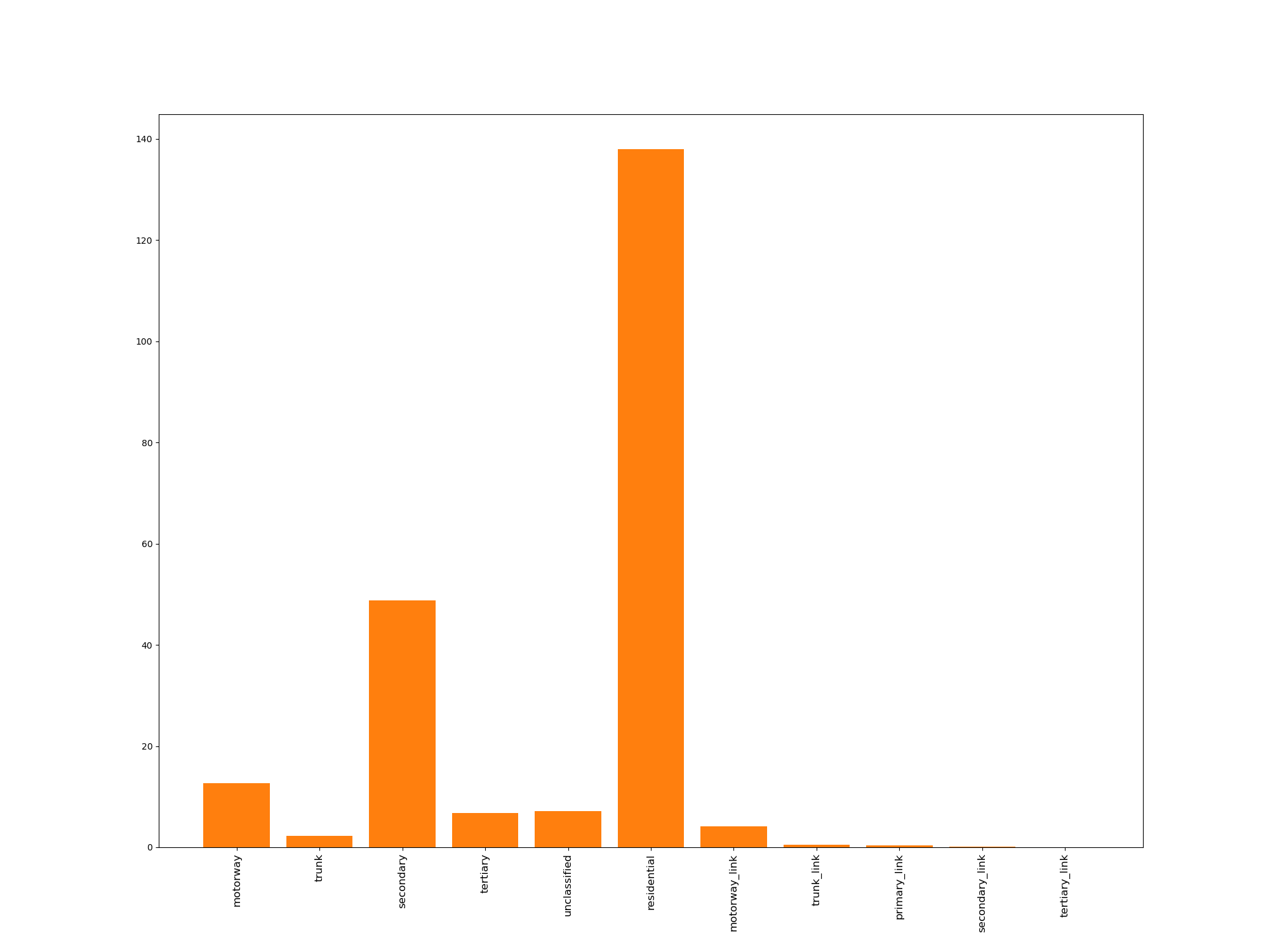}
	\end{minipage}}
 \hfill	
\caption{New York Midtown and Lower Manhattan: road network and its corresponding analysis.}
\label{fig:nycmid}
\end{figure}

Next, we perform analysis on some of the road networks in the hotspots mentioned above. We consider various parameters that are extracted from openstreetmap.com.
We first color-code the road network to provide the visualization of every type of road. Using New York city midtown as an example, Figure~\ref{fig:nycmid} (a) shows the color coded road network, which can help us understand the details of the area and types of road it consists of. To further understand each region in detail, we generate a bar graph representing the percentage of individual types of roads. For example, Figure~\ref{fig:nycmid} (b) and (c) demonstrate the types of roads and their normalized percentage. The bar graph of the normalized results can help us compare road types across different hotspot regions. These graphs give us a chance to understand which types of roads are more prone to accidents. Lastly, as an example, Figure~\ref{fig:nycmid} (d) shows the total length of each type of road in an area.

\begin{figure}[H]
  \subfloat[Road Network in detail]{
	\begin{minipage}[c][1\width]{
	   0.2\textwidth}
	   \centering
	   \includegraphics[width=1.2\textwidth]{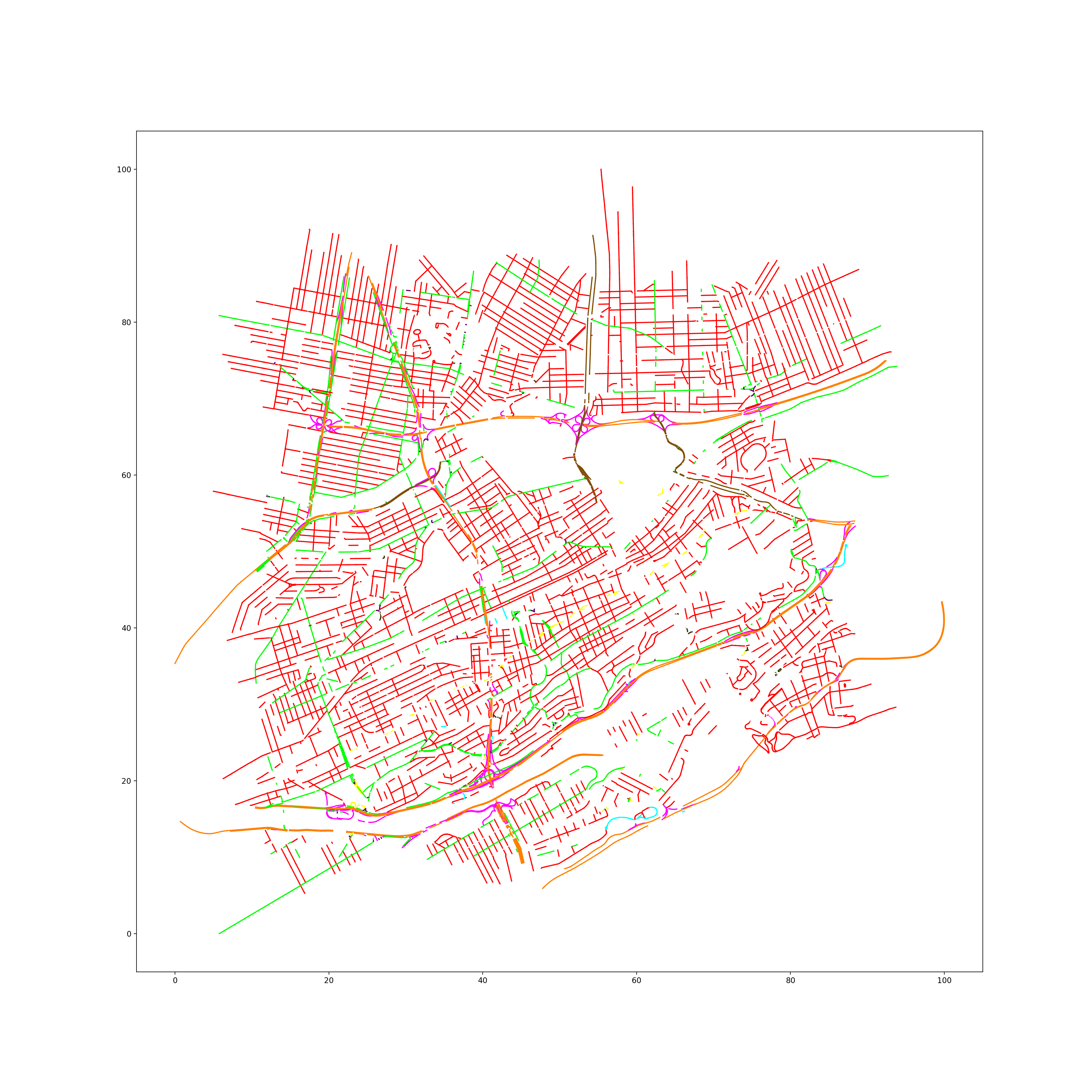}
	\end{minipage}}
 \hfill 	
  \subfloat[Type of highway percentages]{
	\begin{minipage}[c][1\width]{
	   0.2\textwidth}
	   \centering
	   \includegraphics[width=1.2\textwidth]{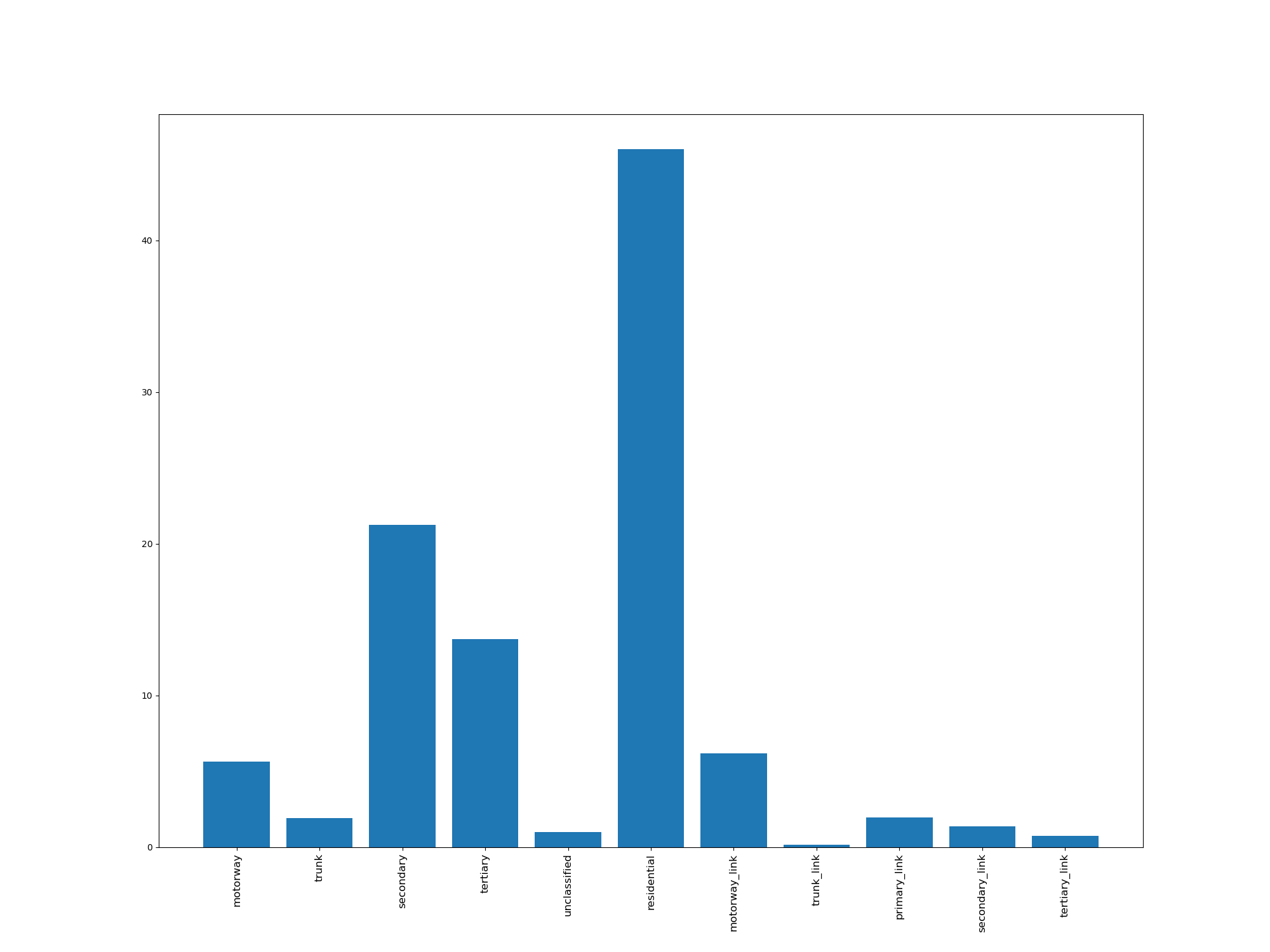}
	\end{minipage}}
 \hfill	
  \subfloat[Type of highway percentage normalized]{
	\begin{minipage}[c][1\width]{
	   0.2\textwidth}
	   \centering
	   \includegraphics[width=1.2\textwidth]{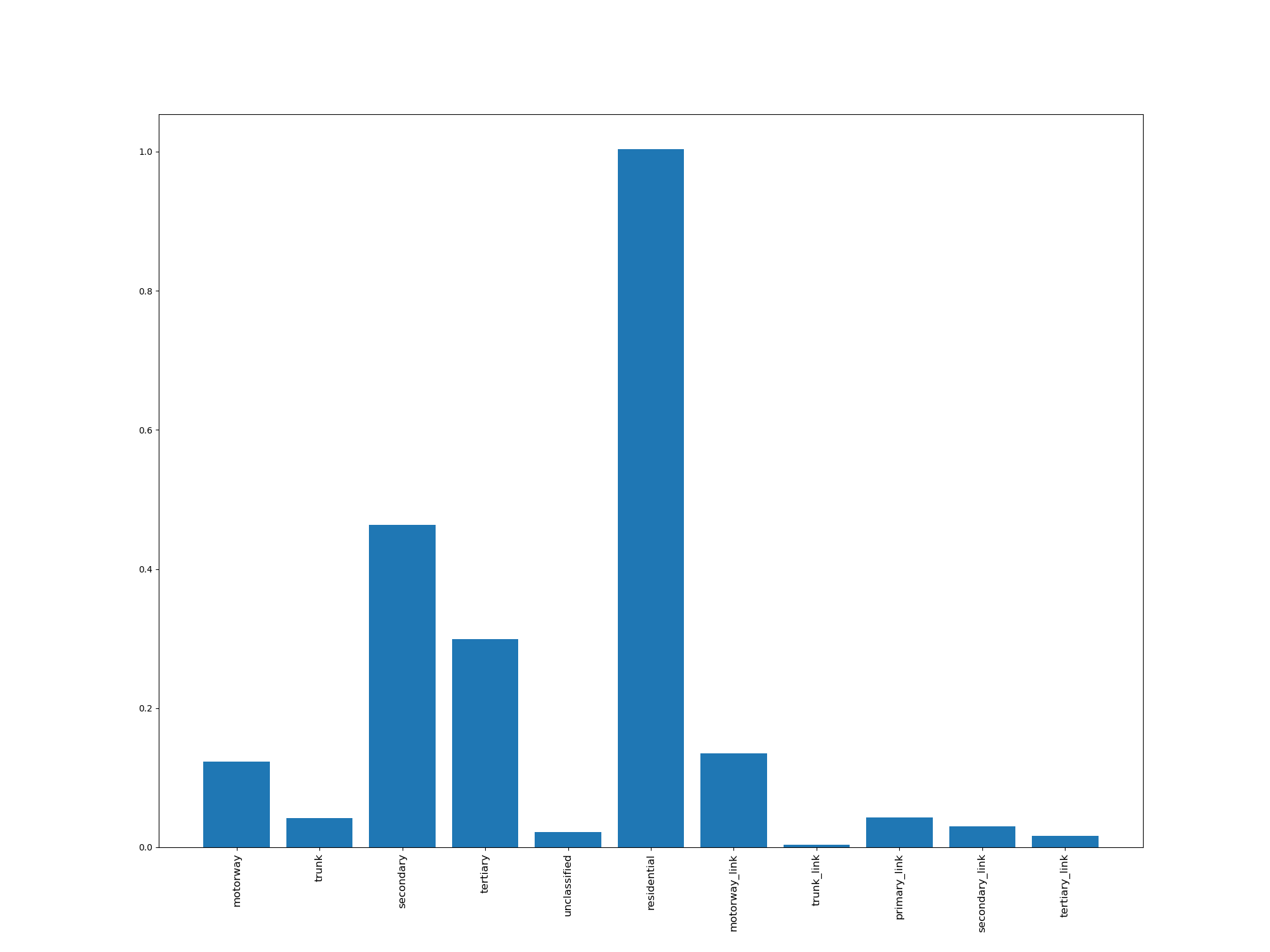}
	\end{minipage}}
 \hfill
  \subfloat[Type of Highways total distance]{
	\begin{minipage}[c][1\width]{
	   0.2\textwidth}
	   \centering
	   \includegraphics[width=1.2\textwidth]{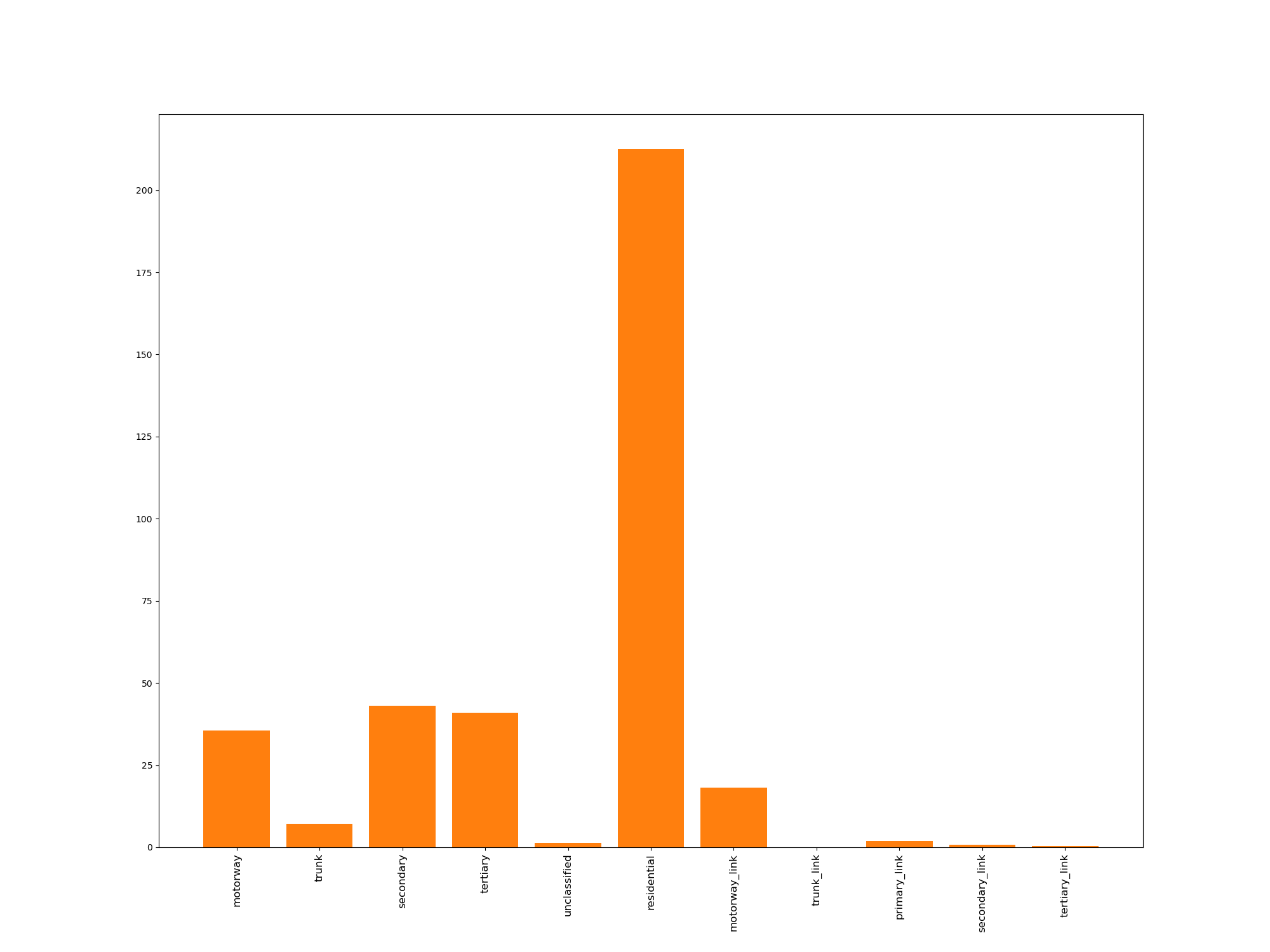}
	\end{minipage}}
 \hfill	
\caption{New York West Bronx: road network and its corresponding analysis.  }
\end{figure}

\begin{figure}[H]
  \subfloat[Road Network in detail]{
	\begin{minipage}[c][1\width]{
	   0.2\textwidth}
	   \centering
	   \includegraphics[width=1.2\textwidth]{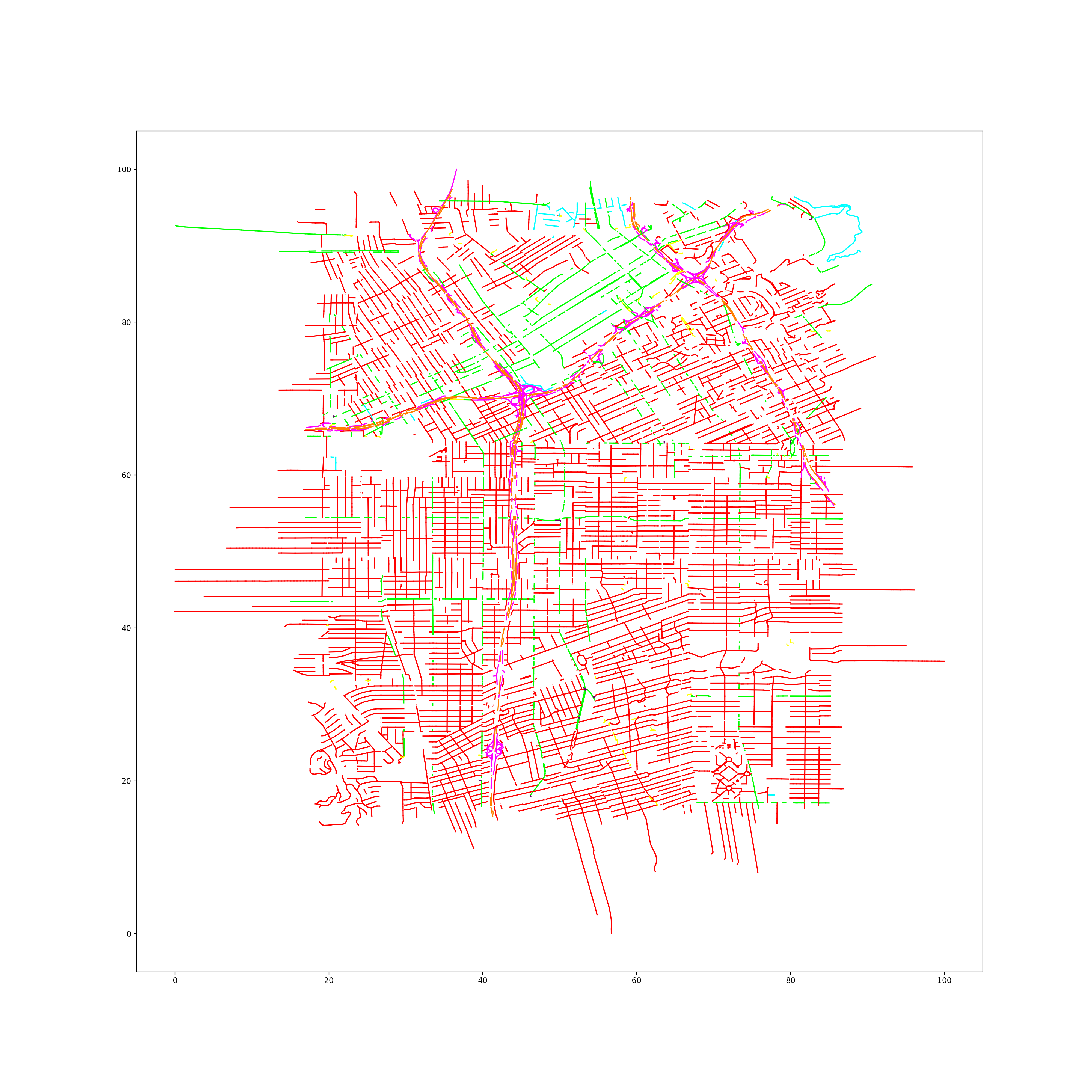}
	\end{minipage}}
 \hfill 	
  \subfloat[Type of highway percentages]{
	\begin{minipage}[c][1\width]{
	   0.2\textwidth}
	   \centering
	   \includegraphics[width=1.2\textwidth]{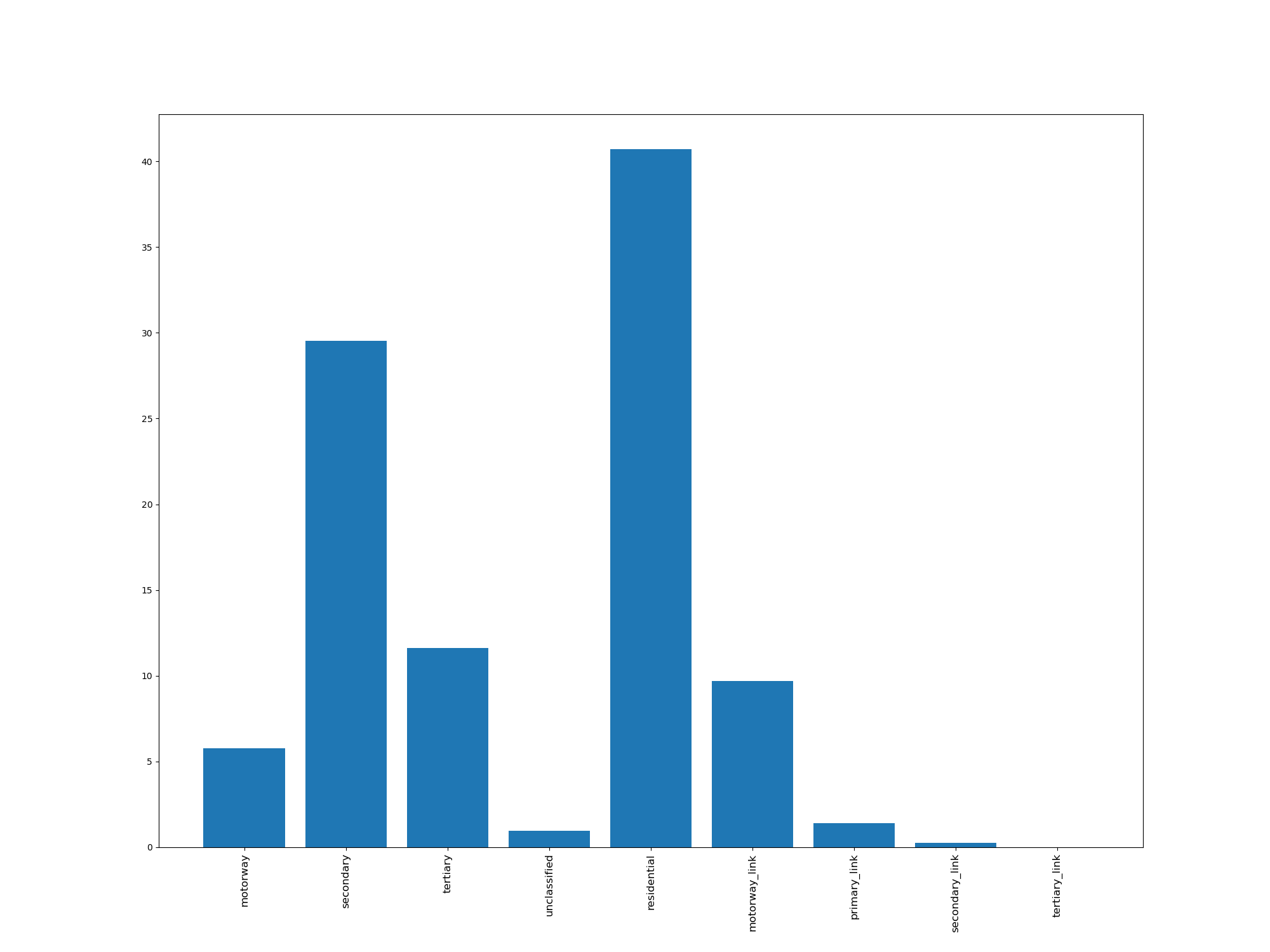}
	\end{minipage}}
 \hfill	
  \subfloat[Type of highway percentage normalized]{
	\begin{minipage}[c][1\width]{
	   0.2\textwidth}
	   \centering
	   \includegraphics[width=1.2\textwidth]{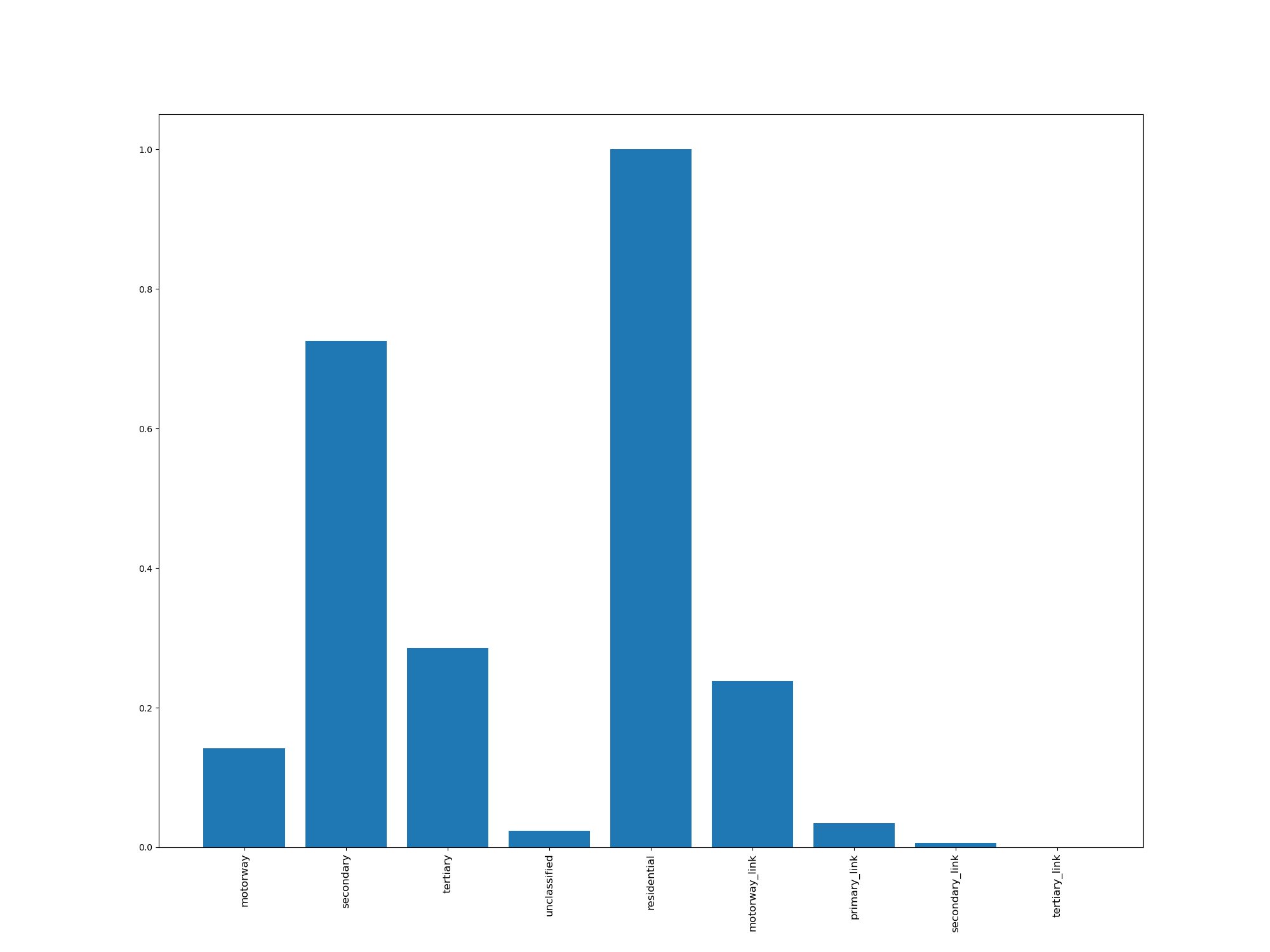}
	\end{minipage}}
 \hfill
  \subfloat[Type of Highways total distance]{
	\begin{minipage}[c][1\width]{
	   0.2\textwidth}
	   \centering
	   \includegraphics[width=1.2\textwidth]{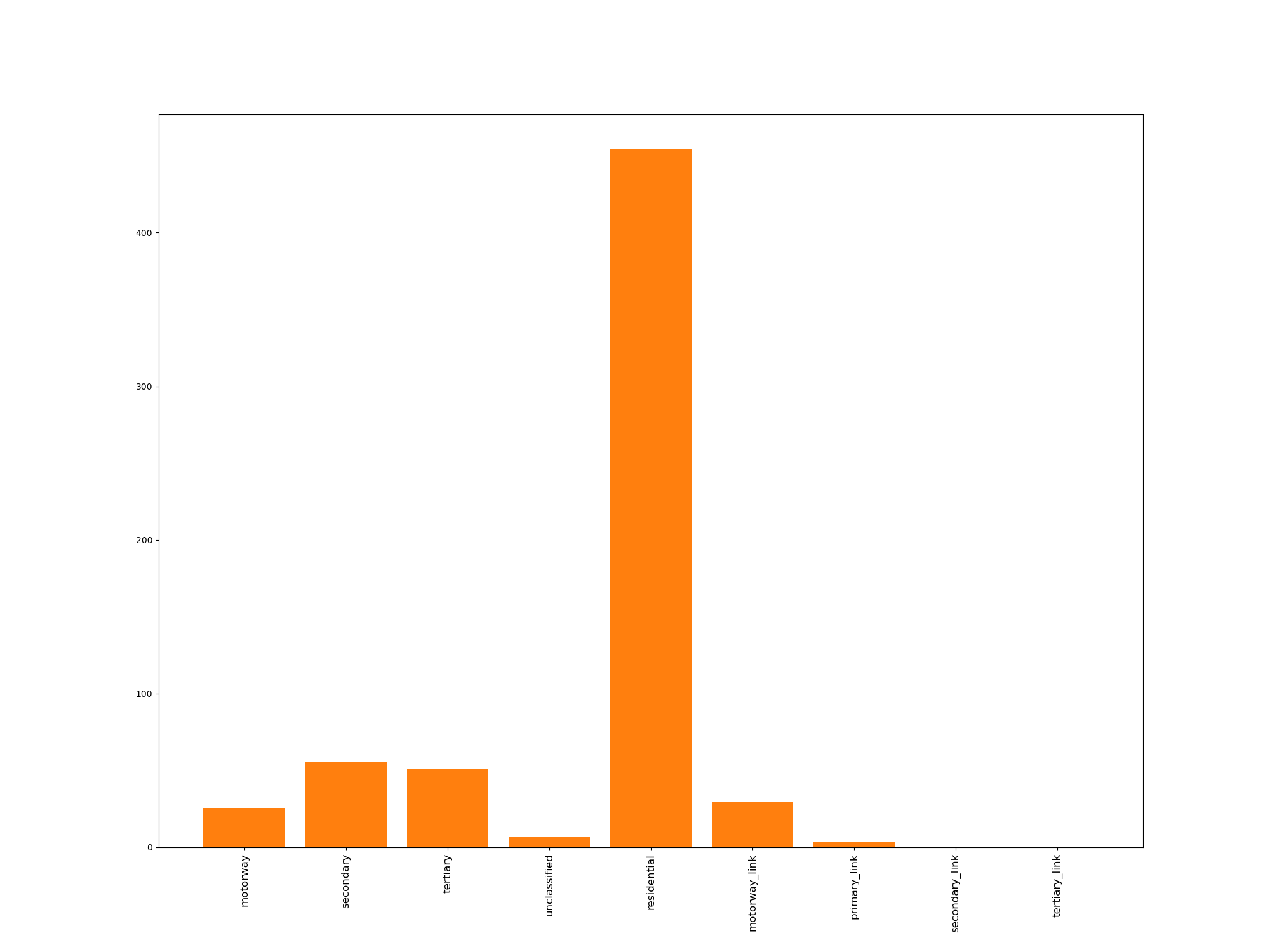}
	\end{minipage}}
 \hfill	
\caption{Northern Los Angeles: road network and its corresponding analysis.  }
\end{figure}

\begin{figure}[H]
  \subfloat[Road Network in detail]{
	\begin{minipage}[c][1\width]{
	   0.2\textwidth}
	   \centering
	   \includegraphics[width=1.2\textwidth]{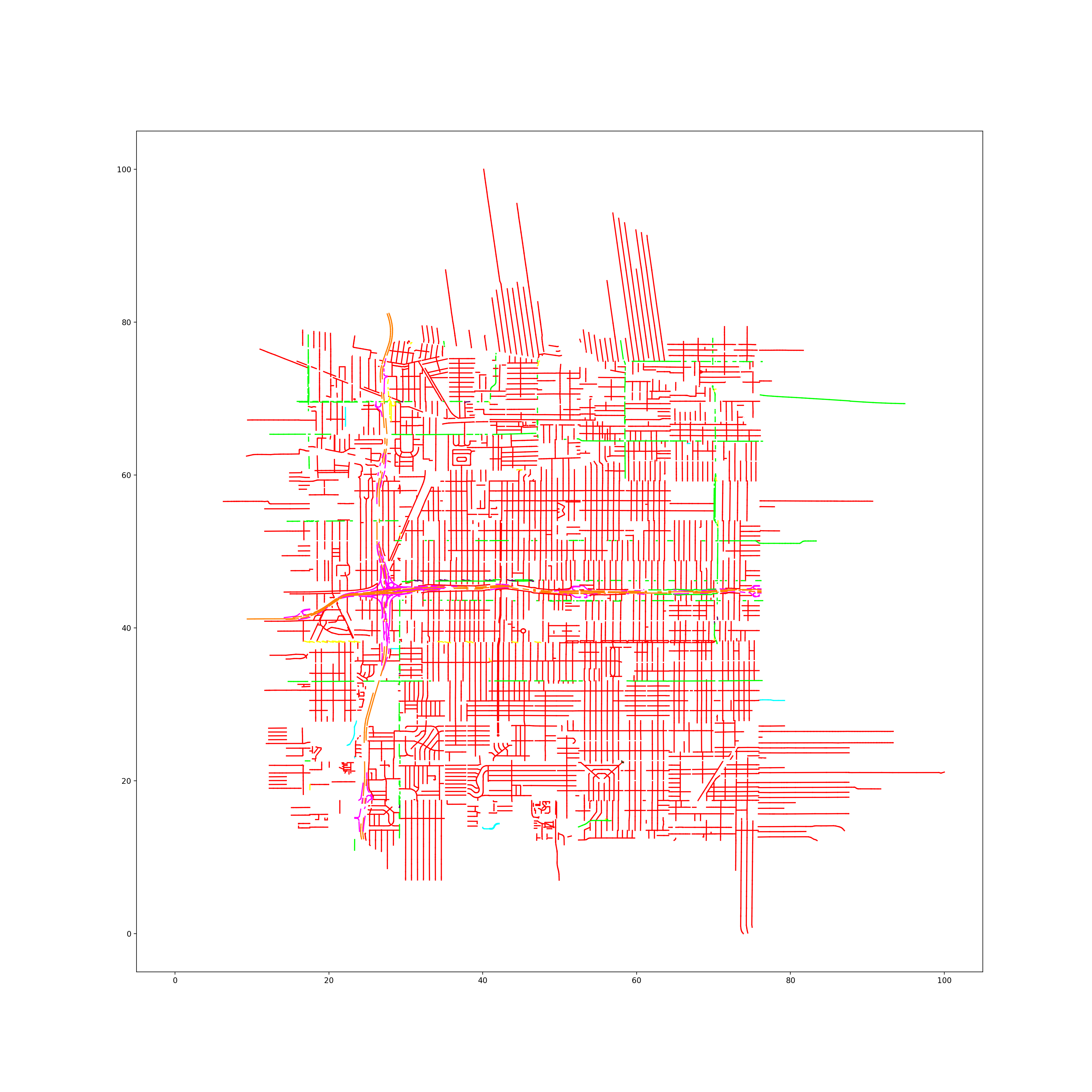}
	\end{minipage}}
 \hfill 	
  \subfloat[Type of highway percentages]{
	\begin{minipage}[c][1\width]{
	   0.2\textwidth}
	   \centering
	   \includegraphics[width=1.2\textwidth]{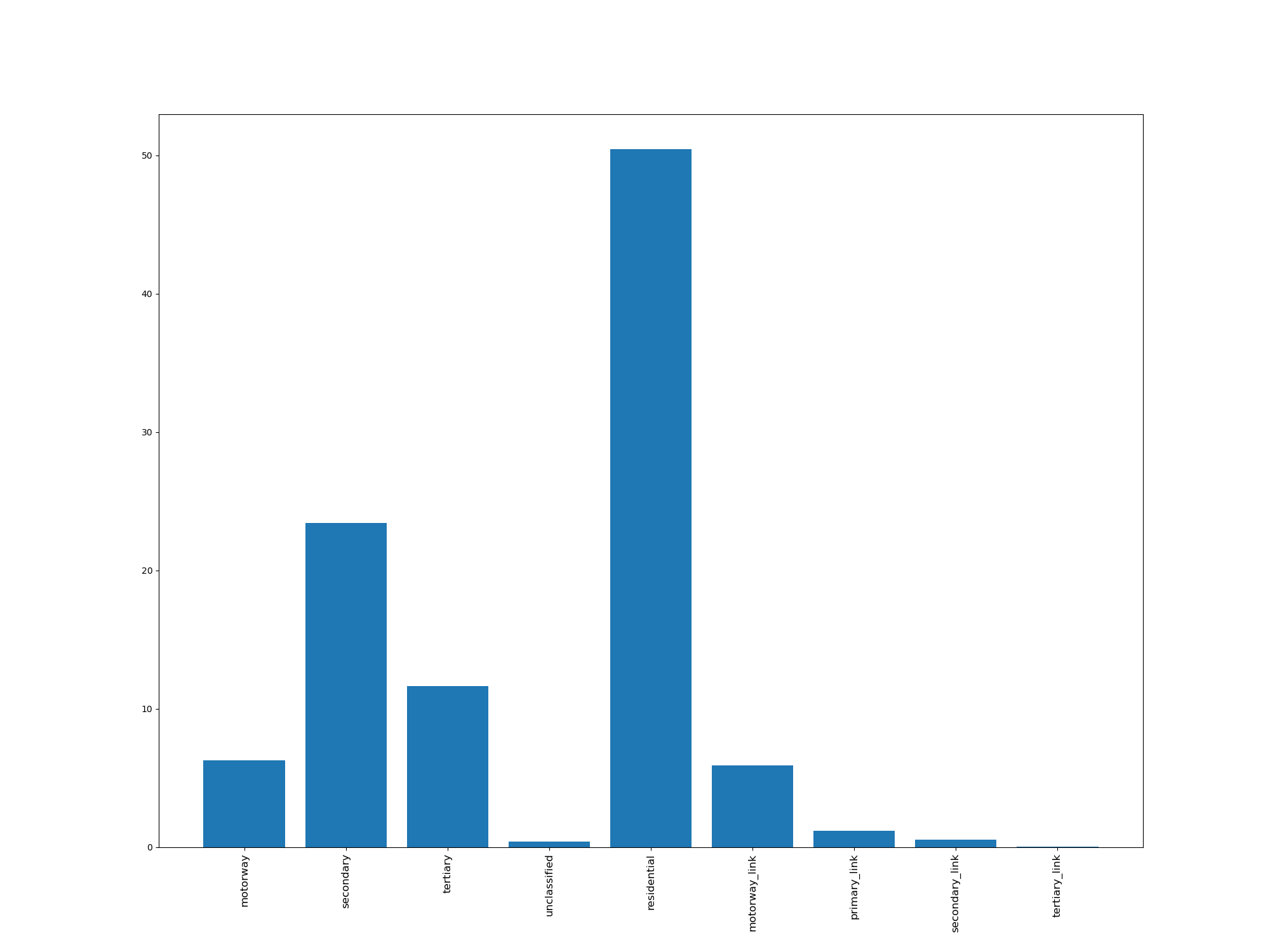}
	\end{minipage}}
 \hfill	
  \subfloat[Type of highway percentage normalized]{
	\begin{minipage}[c][1\width]{
	   0.2\textwidth}
	   \centering
	   \includegraphics[width=1.2\textwidth]{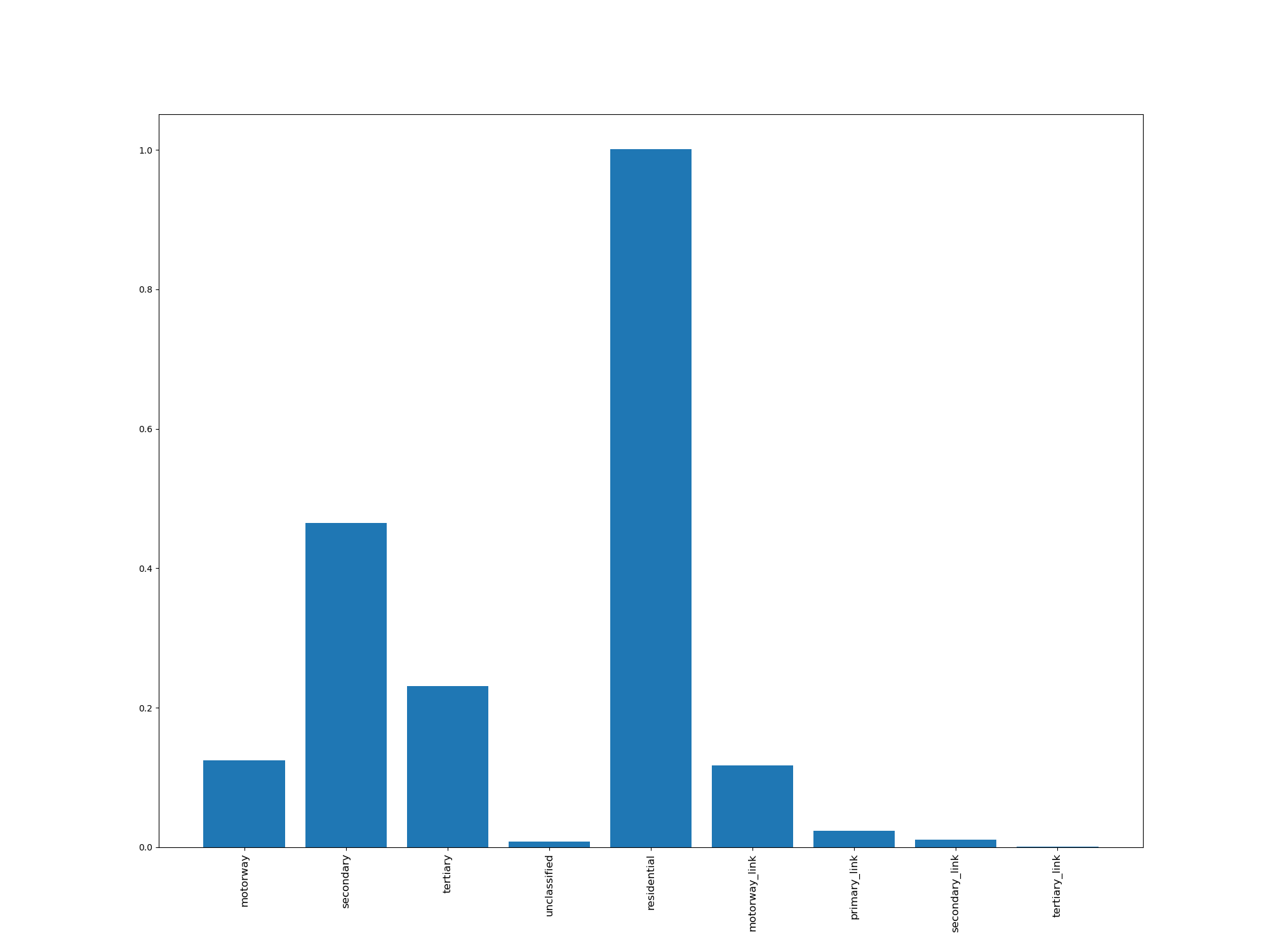}
	\end{minipage}}
 \hfill
  \subfloat[Type of Highways total distance]{
	\begin{minipage}[c][1\width]{
	   0.2\textwidth}
	   \centering
	   \includegraphics[width=1.2\textwidth]{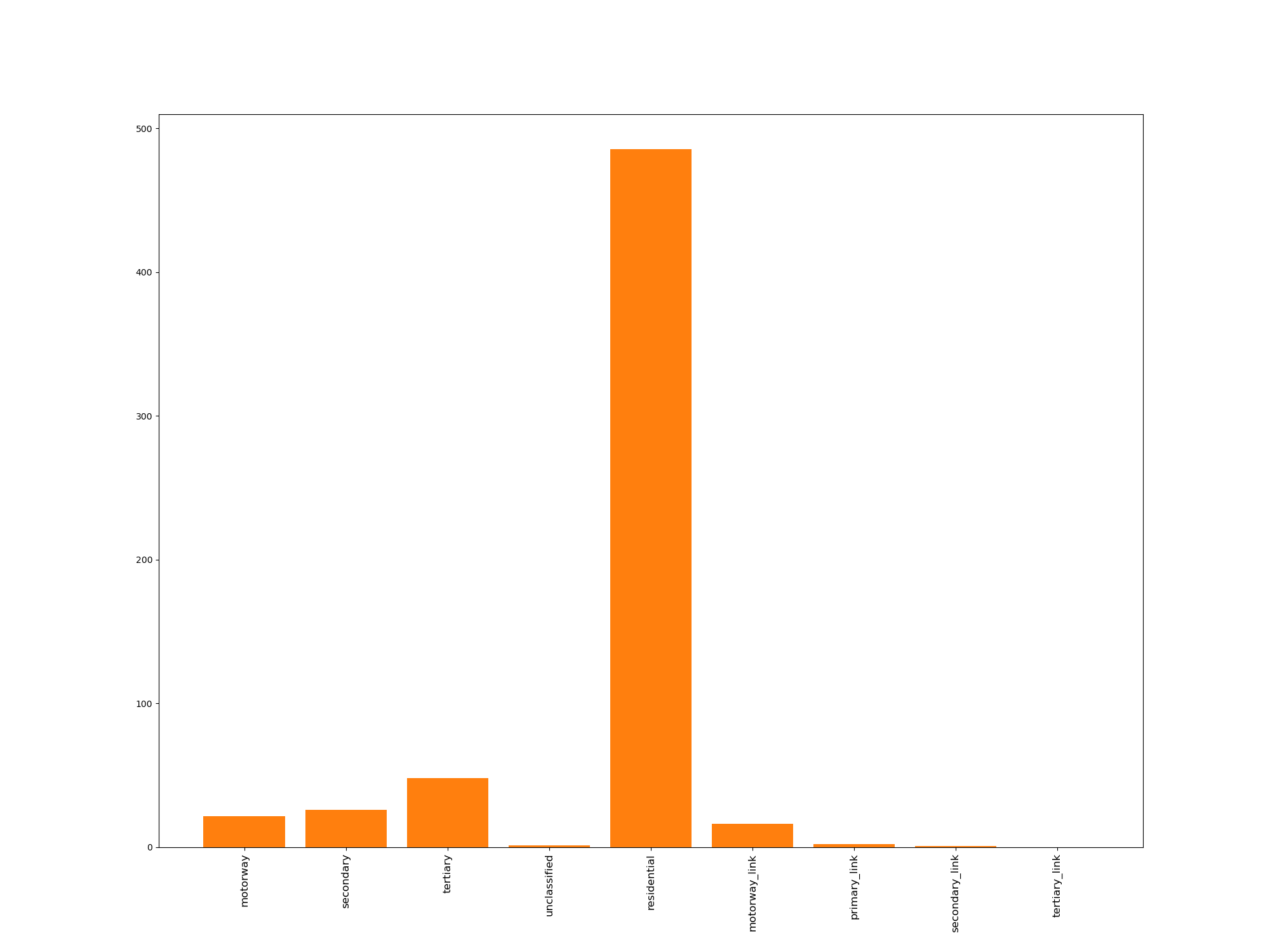}
	\end{minipage}}
 \hfill	
\caption{Southern Los Angeles: road network and its corresponding analysis. }
\end{figure}

\section{Conclusion and Future Work} 
 The influence of COVID-19 on our society and mobility patterns has inspired this study. We have observed that the pandemic has caused a reduction in traffic accident counts and also spatial-temporal shifts of traffic accident hotspots. There are many future directions of this study. First, we want to explore the characteristics of the hotspot regions in more details. Second, we want to explore whether the use of autonomous vehicles can help ease the delivery of essential goods in such critical conditions~\cite{Li2019ADAPS,Chao2020Survey} under realistically simulated, and accurately and robustly estimated traffic conditions~\cite{Wilkie2015Virtual,Li2017CityFlowRecon,Li2017CityEstSparse,Li2018CityEstIter,Lin2019ComSense,Lin2021Attention,Poudel2021Attack}.   



\bibliography{ref}

\begin{thebibliography}{30}
\providecommand{\natexlab}[1]{#1}
\providecommand{\url}[1]{\texttt{#1}}
\expandafter\ifx\csname urlstyle\endcsname\relax
  \providecommand{\doi}[1]{doi: #1}\else
  \providecommand{\doi}{doi: \begingroup \urlstyle{rm}\Url}\fi

\bibitem[Wang et~al.(2021)Wang, Wei, Lin, and Li]{Wang2021Mobility}
S.~Wang, K.~Wei, L.~Lin, and W.~Li.
\newblock Spatial-temporal analysis of {COVID}-19's impact on human mobility:
  the case of the united states.
\newblock In \emph{The 20th and 21st Joint COTA International Conference of
  Transportation Professionals}, 2021.

\bibitem[Lin et~al.(2021)Lin, Shi, and Li]{Lin2021Safety}
L.~Lin, F.~Shi, and W.~Li.
\newblock Assessing inequality, irregularity, and severity regarding road
  traffic safety during covid-19.
\newblock \emph{Scientific Reports}, 11\penalty0 (13147), 2021.
\newblock \doi{https://doi.org/10.1038/s41598-021-91392-z}.

\bibitem[Zhang et~al.(2021)Zhang, Feng, Wu, Xu, Ke, and Dong]{zhang2021effect}
J.~Zhang, B.~Feng, Y.~Wu, P.~Xu, R.~Ke, and N.~Dong.
\newblock The effect of human mobility and control measures on traffic safety
  during covid-19 pandemic.
\newblock \emph{PLoS one}, 16\penalty0 (3):\penalty0 e0243263, 2021.

\bibitem[Xie and Yan(2008)]{xie2008kernel}
Z.~Xie and J.~Yan.
\newblock Kernel density estimation of traffic accidents in a network space.
\newblock \emph{Computers, environment and urban systems}, 32\penalty0
  (5):\penalty0 396--406, 2008.

\bibitem[Kang et~al.(2018)Kang, Cho, and Son]{kang2018spatiotemporal}
Y.~Kang, N.~Cho, and S.~Son.
\newblock Spatiotemporal characteristics of elderly population’s traffic
  accidents in seoul using space-time cube and space-time kernel density
  estimation.
\newblock \emph{PLoS one}, 13\penalty0 (5):\penalty0 e0196845, 2018.

\bibitem[Kim and Yamashita(2007)]{kim2007using}
K.~Kim and E.~Y. Yamashita.
\newblock Using ak-means clustering algorithm to examine patterns of pedestrian
  involved crashes in honolulu, hawaii.
\newblock \emph{Journal of advanced transportation}, 41\penalty0 (1):\penalty0
  69--89, 2007.

\bibitem[Lin et~al.(2014)Lin, Wang, and Sadek]{lin2014data}
L.~Lin, Q.~Wang, and A.~W. Sadek.
\newblock Data mining and complex network algorithms for traffic accident
  analysis.
\newblock \emph{Transportation Research Record}, 2460\penalty0 (1):\penalty0
  128--136, 2014.

\bibitem[Wahl et~al.(2010)Wahl, Islam, Gardner, Marr, Hunt, McSwain, Baker, and
  Duchesne]{wahl2010red}
G.~M. Wahl, T.~Islam, B.~Gardner, A.~B. Marr, J.~P. Hunt, N.~E. McSwain, C.~C.
  Baker, and J.~Duchesne.
\newblock Red light cameras: do they change driver behavior and reduce
  accidents?
\newblock \emph{Journal of Trauma and Acute Care Surgery}, 68\penalty0
  (3):\penalty0 515--518, 2010.

\bibitem[Robson(2001)]{robson2001guide}
L.~S. Robson.
\newblock Guide to evaluating the effectiveness of strategies for preventing
  work injuries; how to show whether a safety invervention really works.
\newblock 2001.

\bibitem[Missoni et~al.(2012)Missoni, Bozic, and Missoni]{missoni2012alcohol}
E.~Missoni, B.~Bozic, and I.~Missoni.
\newblock Alcohol-related road traffic accidents before and after the passing
  of the road traffic safety act in croatia.
\newblock \emph{Collegium antropologicum}, 36\penalty0 (4):\penalty0
  1483--1489, 2012.

\bibitem[Green et~al.(2016)Green, Heywood, and Navarro]{green2016traffic}
C.~P. Green, J.~S. Heywood, and M.~Navarro.
\newblock Traffic accidents and the london congestion charge.
\newblock \emph{Journal of public economics}, 133:\penalty0 11--22, 2016.

\bibitem[Haghpanahan et~al.(2019)Haghpanahan, Lewsey, Mackay, McIntosh, Pell,
  Jones, Fitzgerald, and Robinson]{haghpanahan2019evaluation}
H.~Haghpanahan, J.~Lewsey, D.~F. Mackay, E.~McIntosh, J.~Pell, A.~Jones,
  N.~Fitzgerald, and M.~Robinson.
\newblock An evaluation of the effects of lowering blood alcohol concentration
  limits for drivers on the rates of road traffic accidents and alcohol
  consumption: a natural experiment.
\newblock \emph{The Lancet}, 393\penalty0 (10169):\penalty0 321--329, 2019.

\bibitem[{Transportation Research Board Webinar}(2020)]{trb2020}
{Transportation Research Board Webinar}.
\newblock Traffic trends and safety in a covid-19 world, 2020.
\newblock \url{http://www.trb.org/ElectronicSessions/Blurbs/180648.aspx}, Last
  accessed on 2020-08-28.

\bibitem[{National Safety Council}(2020)]{NSC2020}
{National Safety Council}.
\newblock Motor vehicle fatality rates jump 14\% in march despite quarantines,
  2020.
\newblock
  \url{https://www.nsc.org/in-the-newsroom/motor-vehicle-fatality-rates-jump-14-in-march-despite-quarantines},
  Last accessed on 2020-08-28.

\bibitem[Barnes et~al.(2020)Barnes, Beland, Huh, and Kim]{barnes2020effect}
S.~R. Barnes, L.-P. Beland, J.~Huh, and D.~Kim.
\newblock The effect of covid-19 lockdown on mobility and traffic accidents:
  Evidence from louisiana.
\newblock Technical report, GLO Discussion Paper, 2020.

\bibitem[Truong et~al.(2020)Truong, Oudre, and Vayatis]{truong2020selective}
C.~Truong, L.~Oudre, and N.~Vayatis.
\newblock Selective review of offline change point detection methods.
\newblock \emph{Signal Processing}, 167:\penalty0 107299, 2020.

\bibitem[Goo(2020)]{Google}
{COVID-19}-community mobility reports.
\newblock \url{https://www.google.com/covid19/mobility/}, 2020.
\newblock Accessed: 2020-08-15.

\bibitem[new(2020)]{newyorkcity}
Motor vehicle collisions - crashes, new york city.
\newblock
  \url{https://data.cityofnewyork.us/Public-Safety/Motor-Vehicle-Collisions-Crashes/h9gi-nx95},
  2020.
\newblock Accessed: 2020-08-15.

\bibitem[Lad(2020)]{Ladata}
Traffic collision data from 2010 to present, los angeles city.
\newblock
  \url{https://data.lacity.org/Public-Safety/Traffic-Collision-Data-from-2010-to-Present/d5tf-ez2w},
  2020.
\newblock Accessed: 2020-08-15.

\bibitem[bos(2020)]{bostondata}
Crash query and visualization, boston, ma.
\newblock \url{https://apps.impact.dot.state.ma.us/cdv/}, 2020.
\newblock Accessed: 2020-08-15.

\bibitem[tra(2020)]{trafficnytimes}
The traffic trade-off, ny times, traffic during covid-19.
\newblock
  \url{https://www.nytimes.com/2020/06/04/climate/coronavirus-traffic-air-quality.html},
  2020.
\newblock Accessed: 2020-08-15.

\bibitem[Li et~al.(2019)Li, Wolinski, and Lin]{Li2019ADAPS}
W.~Li, D.~Wolinski, and M.~C. Lin.
\newblock {ADAPS}: Autonomous driving via principled simulations.
\newblock In \emph{IEEE International Conference on Robotics and Automation
  (ICRA)}, pages 7625--7631, 2019.
\newblock \doi{https://doi.org/10.1109/icra.2019.8794239}.

\bibitem[Chao et~al.(2020)Chao, Bi, Li, Mao, Wang, Lin, and
  Deng]{Chao2020Survey}
Q.~Chao, H.~Bi, W.~Li, T.~Mao, Z.~Wang, M.~C. Lin, and Z.~Deng.
\newblock A survey on visual traffic simulation: Models, evaluations, and
  applications in autonomous driving.
\newblock \emph{Computer Graphics Forum}, 39\penalty0 (1):\penalty0 287--308,
  2020.
\newblock \doi{https://doi.org/10.1111/cgf.13803}.

\bibitem[Wilkie et~al.(2015)Wilkie, Sewall, Li, and Lin]{Wilkie2015Virtual}
D.~Wilkie, J.~Sewall, W.~Li, and M.~C. Lin.
\newblock Virtualized traffic at metropolitan scales.
\newblock \emph{Frontiers in Robotics and AI}, 2:\penalty0 11, 2015.
\newblock ISSN 2296-9144.
\newblock \doi{https://doi.org/10.3389/frobt.2015.00011}.

\bibitem[Li et~al.(2017{\natexlab{a}})Li, Wolinski, and
  Lin]{Li2017CityFlowRecon}
W.~Li, D.~Wolinski, and M.~C. Lin.
\newblock City-scale traffic animation using statistical learning and
  metamodel-based optimization.
\newblock \emph{ACM Trans. Graph.}, 36\penalty0 (6):\penalty0 200:1--200:12,
  Nov. 2017{\natexlab{a}}.
\newblock \doi{https://doi.org/10.1145/3130800.3130847}.

\bibitem[Li et~al.(2017{\natexlab{b}})Li, Nie, Wilkie, and
  Lin]{Li2017CityEstSparse}
W.~Li, D.~Nie, D.~Wilkie, and M.~C. Lin.
\newblock Citywide estimation of traffic dynamics via sparse {GPS} traces.
\newblock \emph{IEEE Intelligent Transportation Systems Magazine}, 9\penalty0
  (3):\penalty0 100--113, 2017{\natexlab{b}}.
\newblock ISSN 1939-1390.
\newblock \doi{https://doi.org/10.1109/mits.2017.2709804}.

\bibitem[Li et~al.(2018)Li, Jiang, Chen, and Lin]{Li2018CityEstIter}
W.~Li, M.~Jiang, Y.~Chen, and M.~C. Lin.
\newblock Estimating urban traffic states using iterative refinement and
  wardrop equilibria.
\newblock \emph{IET Intelligent Transport Systems}, 12\penalty0 (8):\penalty0
  875--883, 2018.
\newblock \doi{https://doi.org/10.1049/iet-its.2018.0007}.

\bibitem[Lin et~al.(2019)Lin, Li, and Peeta]{Lin2019ComSense}
L.~Lin, W.~Li, and S.~Peeta.
\newblock Efficient data collection and accurate travel time estimation in a
  connected vehicle environment via real-time compressive sensing.
\newblock \emph{Journal of Big Data Analytics in Transportation}, 1\penalty0
  (2):\penalty0 95--107, 2019.
\newblock \doi{https://doi.org/10.1007/s42421-019-00009-5}.

\bibitem[Lin et~al.(2021)Lin, Li, Bi, and Qin]{Lin2021Attention}
L.~Lin, W.~Li, H.~Bi, and L.~Qin.
\newblock Vehicle trajectory prediction using {LSTM}s with spatial-temporal
  attention mechanisms.
\newblock \emph{IEEE Intelligent Transportation Systems Magazine}, 2021.

\bibitem[Poudel and Li(2021)]{Poudel2021Attack}
B.~Poudel and W.~Li.
\newblock Black-box adversarial attacks on network-wide multi-step traffic
  state prediction models.
\newblock In \emph{IEEE International Conference on Intelligent Transportation
  Systems (ITSC)}, 2021.

\end{thebibliography}

\end{document}